\documentclass[12pt]{elsarticle}

\usepackage{xcolor}
\usepackage{multirow}
\usepackage{booktabs}
\usepackage{graphicx}
\usepackage[labelsep=period]{caption}

\usepackage{colortbl} 
\usepackage{placeins}  
\usepackage[margin=1in]{geometry}
\usepackage{longtable}
\usepackage{caption} 
\usepackage[left]{lineno}
\usepackage{float}
\usepackage{afterpage}
\usepackage[most]{tcolorbox}
\usepackage{bookmark}
\usepackage{makecell}
\usepackage{setspace}
\usepackage{enumitem}
\usepackage{soul} 

\usepackage{amssymb}
\usepackage{amsmath}
\usepackage{xcolor}
\usepackage{url}

\definecolor{softblue}{RGB}{30, 60, 150}

\journal{Journal of Biomedical Informatics}

\begin{document}
\onehalfspacing

\begin{frontmatter}

\title{Vision-Language Model-Based Semantic-Guided Imaging Biomarker for Lung Nodule Malignancy Prediction}

\author[label1]{Luoting Zhuang} 
\author[label1]{Seyed Mohammad Hossein Tabatabaei}
\author[label2，label3]{Ramin Salehi-Rad}
\author[label2，label3]{Linh M. Tran}
\author[label1]{Denise R. Aberle}
\author[label1]{Ashley E. Prosper}
\author[label1]{William Hsu\corref{cor1}}

\cortext[cor1]{Corresponding author. Email: wshu@mednet.ucla.edu}

\affiliation[label1]{organization={Medical \& Imaging Informatics, Department of Radiological Sciences, David Geffen School of Medicine at UCLA},
            city={Los Angeles},
            postcode={90095}, 
            state={CA},
            country={USA}}

\affiliation[label2]{organization={Department of Medicine, Division of Pulmonology and Critical Care, David Geffen School of Medicine at UCLA},
            city={Los Angeles},
            postcode={90095}, 
            state={CA},
            country={USA}}

\affiliation[label3]{organization={VA Greater Los Angeles Healthcare System},
            city={Los Angeles},
            postcode={90073}, 
            state={CA},
            country={USA}}

\begin{abstract}
\textbf{Objective:} Machine learning models have utilized semantic features, deep features, or both to assess lung nodule malignancy. However, their reliance on manual annotation during inference, limited interpretability, and sensitivity to imaging variations hinder their application in real-world clinical settings. Thus, this research aims to integrate semantic features derived from radiologists’ assessments of nodules, guiding the model to learn clinically relevant, robust, and explainable imaging features for predicting lung cancer. 

\noindent\textbf{Methods:} We obtained 938 low-dose CT scans from the National Lung Screening Trial (NLST) with 1,261 nodules and semantic features. Additionally, the Lung Image Database Consortium dataset contains 1,018 CT scans, with 2,625 lesions annotated for nodule characteristics. Three external datasets were obtained from UCLA Health, the LUNGx Challenge, and the Duke Lung Cancer Screening. For imaging input, we obtained 2D nodule slices in nine directions from $50\times50\times50$ mm nodule crop. We converted structured semantic features into sentences using Gemini. We fine-tuned a pretrained Contrastive Language-Image Pretraining (CLIP) model with a parameter-efficient fine-tuning approach to align imaging and semantic text features and predict the one-year lung cancer diagnosis.

\noindent\textbf{Results:} Our model outperformed the state-of-the-art (SOTA) models in the NLST test set with an AUROC of 0.901 and AUPRC of 0.776. It also showed robust results in external datasets. Using CLIP, we also obtained predictions on semantic features through zero-shot inference, such as nodule margin (AUROC: 0.807), nodule consistency (0.812), and pleural attachment (0.840).

\noindent\textbf{Conclusion:} By incorporating semantic features into the vision-language model, our approach surpasses the SOTA models in predicting lung cancer from CT scans collected from diverse clinical settings. It provides explainable outputs, aiding clinicians in comprehending the underlying meaning of model predictions. The code is available at \url{https://github.com/luotingzhuang/CLIP_nodule}.

\end{abstract}

\begin{graphicalabstract}
\includegraphics[width=\textwidth]{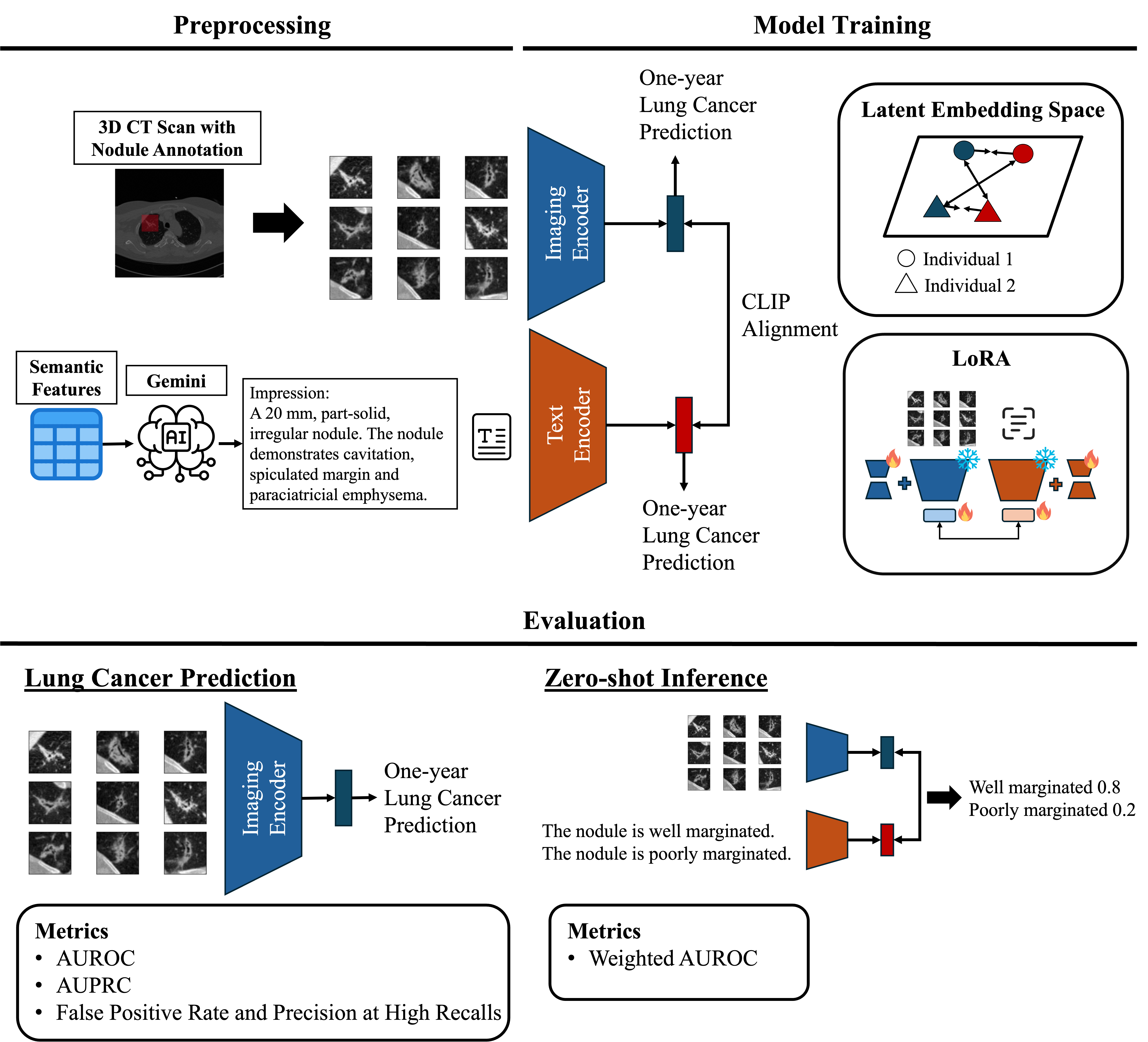}
\end{graphicalabstract}


\end{frontmatter}

\section{Introduction}
\label{intro}
Lung cancer remains the leading cause of cancer-related deaths \cite{barta2019global,cruz2011lung}. Computed tomography (CT) has demonstrated effectiveness in detecting lung cancer and decreasing cancer-related mortality during lung cancer screening and also in routine healthcare settings \cite{nlst,nelson}. However, radiologists are experiencing burnout due to the growing number of detected nodules resulting from the increasing application of CT. Computer-aided diagnostic systems have been proposed to alleviate the workload of radiologists by providing automatic and accurate predictions of lung cancer based on patient data.

Traditional machine learning models in lung nodule risk assessment focus on semantic features, deep features, or both. Semantic features refer to the descriptive terms radiologists use to characterize regions of interest, such as the shape, margin, and vascularity. Models trained on semantic features are easier to interpret, but they also introduce difficulties when scaled up for clinical use, as they require manual annotation from radiologists \cite{wu2019comparison, bashir2019non}. Recently, deep learning (DL) has become a powerful tool due to its strong capability of extracting complex features from CT without any manual input \cite{sybil,venkadesh}. Such imaging-based models still encounter numerous obstacles. For instance, imaging features are sensitive to variations in the acquisition and reconstruction parameters of CT scans, including dose levels, slice thicknesses, and reconstruction kernels \cite{zhuang2024exploring,emaminejad2021reproducibility}. Moreover, research has indicated that DL models learn shortcuts, which are characteristics highly correlated with outcomes but lack clinical significance \cite{geirhos2020shortcut,degrave2021ai}. This phenomenon could adversely impact the reproducibility of features and the generalization of models. Additionally, the lack of explainability remains one of the significant challenges related to deep features. Heatmaps or attribution maps generated by explainability methods, such as GradCAM \cite{gradcam}, can only indicate the regions on which the model focuses, but do not reveal the specific features utilized.

Combining these features has become more popular for enhancing lung cancer prediction performance. However, directly merging two features often reduces applicability in clinical settings, as manual annotations are still required. Alternative approaches tackle this problem through co-learning and multi-task learning \cite{shen2019explainable, gao2022reducing, liu2019multi}, but these models face challenges regarding explainability or training difficulties when dealing with numerous semantic features. An emerging approach in vision-language models (VLM), Contrastive Language-Image Pretraining (CLIP) \cite{clip}, fills the gap by learning the alignment between imaging features and descriptive text. Guided by text, these models allow deep visual features to capture a richer and more robust representation. Therefore, in this study, we have made the following key contributions:
\begin{enumerate}[itemsep=2pt, parsep=1pt]
    \item We employed CLIP to incorporate semantic features, directing the model in obtaining clinically significant and robust features for predicting lung cancer. 
    \item We curated a unique dataset comprising CT scans and their corresponding semantic features, which were annotated by radiologists. The semantic features encompass characteristics of both nodules and their surrounding environment. 
    \item We benchmarked our CLIP model against several state-of-the-art (SOTA) lung cancer prediction models using a comprehensive set of metrics. Our model demonstrated better and more robust results in external datasets collected from different clinical scenarios, such as screening and routine clinical care, and different types of CT scans, such as low-dose CT (LDCT), diagnostic CT, and contrast-enhanced CT (CECT). 
    \item We explored the zero-shot inference, a feature of CLIP-based models that generates semantic features despite not being explicitly trained to do so.
    \item Several adjustments were introduced to accommodate CLIP fine-tuning on 3D images with limited data. By combining these strategies with extensive domain knowledge, our model shows comparable and better performance than imaging-based models trained on tens of thousands of cases using an order of magnitude fewer cases.
\end{enumerate}

\begin{table}[H]
\centering
\renewcommand{\arraystretch}{1.4}
\resizebox{\textwidth}{!}{\begin{tabular}{p{3cm}p{13cm}}
\toprule
\multicolumn{2}{l}{\textbf{Statement of Significance}} \\
\midrule
Problem or Issue & Imaging-based lung nodule malignancy prediction models lack explainability and tend to learn shortcuts, limiting their generalizability and applicability in diverse clinical settings.  \\
What is Already Known & One of the VLMs, CLIP, aligns image and text features in a shared space to integrate visual and textual information effectively. \\
What this Paper Adds & We utilize CLIP to incorporate both internal and external semantic features to guide the imaging model to learn clinically relevant and reliable imaging biomarkers, aiming to improve lung cancer classification across diverse patient populations and CT acquisition protocols. This model offers explainability through zero-shot inference, helping radiologists and end users understand model predictions with semantic features.  \\

\bottomrule
\end{tabular}
}
\end{table}

\section{Related Work}
\label{relatedwork}
\subsection{Imaging-based Lung Cancer Prediction Model}
DL models trained on CT scans have been made available online for potential implementation in real-world clinical settings. Sybil \cite{sybil} employs ResNet18 to predict lung cancer risk scores for up to six years using LDCT scans. Sybil was trained using the National Lung Screening Trial (NLST) dataset, with around 15,000 CT scans, and has consistently demonstrated robust performance during external validation using screening scans from two distinct sites. In another study, Venkadesh et al. \cite{venkadesh} developed a model to predict the malignancy risk of pulmonary nodules by integrating ResNet18 for 2D nodule crops and Inception-V1 for 3D nodule crops. This model was trained using 16,077 nodules from the NLST dataset and has shown robust performance in external cancer screening datasets.

Although these models have shown effectiveness in predicting lung cancer, they still lack explainability. It is unclear what features have been captured from the image. Moreover, as these models were trained on NLST data and only evaluated on screening cohorts, they can overfit to LDCT scans and have a limited scope of clinical application, such as in diagnostic CT and CECT. Studies have also shown that the risk scores generated from Sybil are sensitive to the CT reconstruction parameters, such as slice thickness and kernel \cite{zhuang2024exploring}.

\subsection{Imaging Feature Learning Guided by Semantic Features}
Several studies have also investigated the incorporation of clinical or semantic features to improve the accuracy of lung cancer prediction. These models can capture the relationship between imaging and semantic data, making the resulting embeddings more clinically meaningful and interpretable. However, unlike medical imaging, which is more readily available, semantic features often require labor-intensive annotation by radiologists. The existing approach of directly fusing two modalities can limit clinical utility and increase radiologists' workload. Several studies attempted to address the issue. For example, DeepLungIPN \cite{gao2022reducing} is a co-learning model trained on an in-house dataset of 1,284 cases, integrating both imaging features and clinical features, which consist of demographics and nodule characteristics. Imaging features were obtained from the top five most likely nodules based on a nodule detection algorithm. They implemented three prediction branches, each for the imaging feature, clinical tabular feature, and the fusion of the two. However, the model cannot generate explanations of its outputs, despite jointly modeling information across two modalities. 

Moreover, a hierarchical semantic convolutional neural network \cite{shen2019explainable} was developed using a multi-task learning framework. It simultaneously performs a low-level task of predicting semantic features and a high-level task of assessing nodule malignancy. The model combines embeddings from the low-level semantic predictions with global convolutional representations to generate the final malignancy assessment. This architecture is inherently interpretable, as it provides explicit predictions for semantic features through its low-level branches. However, challenges arise when a large number of semantic features are involved. Conflicting tasks can lead to inconsistent gradient updates in the shared layers, ultimately degrading overall model performance.

\subsection{Vision-Language Model}
VLMs have recently gained popularity, employing DL to understand visual and linguistic information simultaneously. Generally, joint representations learning involves mapping features from different modalities into a shared latent space. One of the VLMs is CLIP \cite{clip}, which aligns natural images with their corresponding captions. It has exhibited exceptional performance across a variety of downstream tasks. This framework has also been successfully applied in the medical field, as clinicians generally write reports linked to various medical modalities, such as echocardiograms \cite{christensen2024vision}, chest X-rays \cite{zhang2023knowledge}, histopathology images \cite{huang2023visual}, and CT scans \cite{hamamci2024developing}. 

The CLIP framework can also be suitable for predicting lung cancer by aligning nodule-specific imaging features with semantic features. By projecting both feature types from the same patient into a shared embedding space, the model enables image features to incorporate semantic context. Additionally, the model offers greater explainability, as it can perform zero-shot inference to evaluate how closely an image aligns with a given semantic feature. However, several questions arise when training such a contrastive learning model on medical imaging and semantic features. First, the success of CLIP relies on a large dataset. Although medical imaging data and reports can be collected with relative ease, it is challenging to obtain detailed characterizations for specific regions like nodules. Therefore, training a VLM model from scratch is impractical, and it is necessary to use a pretrained CLIP model. However, issues emerge when 3D medical images are input into a pretrained CLIP model, which is only compatible with 2D images. Second, semantic features are typically presented in tabular format and may contain missing values. Such tabular data can also restrict the applicability of the pretrained text encoder from the CLIP model, which already maintains some degree of alignment between images and text.

CLIP-Lung \cite{cliplung} was introduced to utilize CLIP for learning generalized visual representations from semantic features using the Lung Image Database Consortium (LIDC) dataset. However, one of the critical limitations of the study is the lack of investigation into one of CLIP’s most distinctive capabilities, zero-shot inference. While CLIP is not trained to predict semantic features, zero-shot inference identifies semantic features most similar to a given imaging feature embedding. Leveraging this capability could improve model explainability and enhance radiologists’ trust in the system. Second, the authors did not perform a benchmark to evaluate the model against SOTA lung cancer risk prediction models across datasets with different properties. Additionally, the model made predictions of risk score estimations from radiologists instead of biopsy-confirmed results. Furthermore, the model was trained on a limited number of semantic features that do not fully characterize the properties of the nodule and the surrounding areas.

\section{Methods}
\label{methods}

\subsection{Datasets}
\subsubsection{Training and Test Data}

\begin{table}[!t]
\centering
\small
\caption{\textbf{Datasets.} Datasets utilized in this study are summarized with respect to cohort size, proportion of lung cancer cases, imaging modality, and data source. It is important to note that the biopsy-confirmed diagnoses for lung cancer in the LIDC dataset are incomplete. Abbreviations: NLST --- National Lung Screening Trial; LIDC --- Lung Image Database Consortium; DLCS --- Duke Lung Cancer Screening; LDCT --- Low-dose Computed Tomography; CECT --- Contrast-enhanced Computed Tomography. Symbols: $\cup$ - Union; $\cap$ - Intersection}
\resizebox{\textwidth}{!}{
\begin{tabular}{lcccc}
\toprule
\textbf{Dataset} & \textbf{\# Cases (\% positive)} & \textbf{\# Nodules} &\textbf{CT Type} &\textbf{Patient Cohort}\\
\midrule
\multicolumn{5}{c}{\textbf{Training Datasets}} \\
\midrule

NLST & 938 (23\%) &1,261& LDCT &Screening \\
LIDC & 1,018 (9\%) &2,625& LDCT \& Diagnostic CT & Screening $\cup$ Incidental  \\

\midrule
\multicolumn{5}{c}{\textbf{External Datasets}} \\
\midrule

LUNGx Challenge& 70 (50\%)& 83 &Diagnostic CT \& CECT &Incidental \\
UCLA Health&51 (55\%)& 52 &LDCT \& Diagnostic CT & Incidental $\cap$ Never-smoker\\
DLCS &856 (11\%)& 1,388 &LDCT&Screening \\
\bottomrule
\end{tabular}
}
\label{tab:datasets}
\end{table}
The training data were collected from two publicly available datasets (Table \ref{tab:datasets}). First, we obtained 938 LDCT scans containing at least one nodule from the NLST \cite{nlst,tcianlst}. This is a unique cohort for which thorough annotations were performed by fellowship-trained thoracic radiologists at UCLA Health. They annotated the nodule's location and curated a holistic set of 19 semantic features, listed in Table \ref{tab:semantic}, for a total of 1,261 nodules \cite{aberle2025keynote}. These features encompass general features (e.g., shape, margin, and consistency), internal characteristics (e.g., necrosis and cyst-like spaces), and external features (e.g., vascular convergence and emphysema). Additionally, the LIDC dataset contains both LDCT and diagnostic CT scans from 1,018 cases, with 2,625 lesions annotated for nodule characteristics, including sphericity, lobulation, consistency, internal structure, margin, and spiculation \cite{lidc,tcialidc}. Since multiple radiologists annotated the nodules in the study, and their annotations varied, we used a majority vote at the pixel level for the nodule annotations and took the median of the semantic feature scores.

\subsubsection{External Data}
\label{externaldata}
We collected three external datasets, each representing distinct patient cohorts and CT imaging types. First, we obtained 51 diagnostic chest CT scans from a subset of patients at UCLA Health who underwent percutaneous CT-guided lung biopsies. These patients self-identified as never-smokers and were not eligible for lung cancer screening. Despite this, they are considered higher-risk patients due to the incidental detection of nodules that warranted biopsy. Fellowship-trained radiologists annotated the locations of the most suspicious nodules. Second, we utilized the publicly available LUNGx Challenge dataset, which includes 70 diagnostic CT and CECT scans with 83 annotated nodules \cite{lungx,tcialungx}. Lastly, the Duke Lung Cancer Screening (DLCS) dataset comprises 1,613 LDCT scans with 2,487 nodules, collected as part of a lung cancer screening program. Nodule detection was performed using a DL-based algorithm, and some nodules were further reviewed by medical students and radiologists \cite{dlcs}. We excluded individuals with less than one year of follow-up, resulting in a final dataset of 856 patients and 1,388 nodules.

These three datasets also differ in their lung cancer prevalence. The LUNGx and UCLA datasets have balanced distributions, with 50\% (35 cases) and 55\% (28 cases) of nodules diagnosed as malignant tumors, respectively. In contrast, the DLCS dataset has a lower malignancy rate, with only 11\% (94 cases) confirmed as lung cancer, reflective of a screening population.

\begin{figure}[!t]
\centerline{\includegraphics[width=\textwidth]{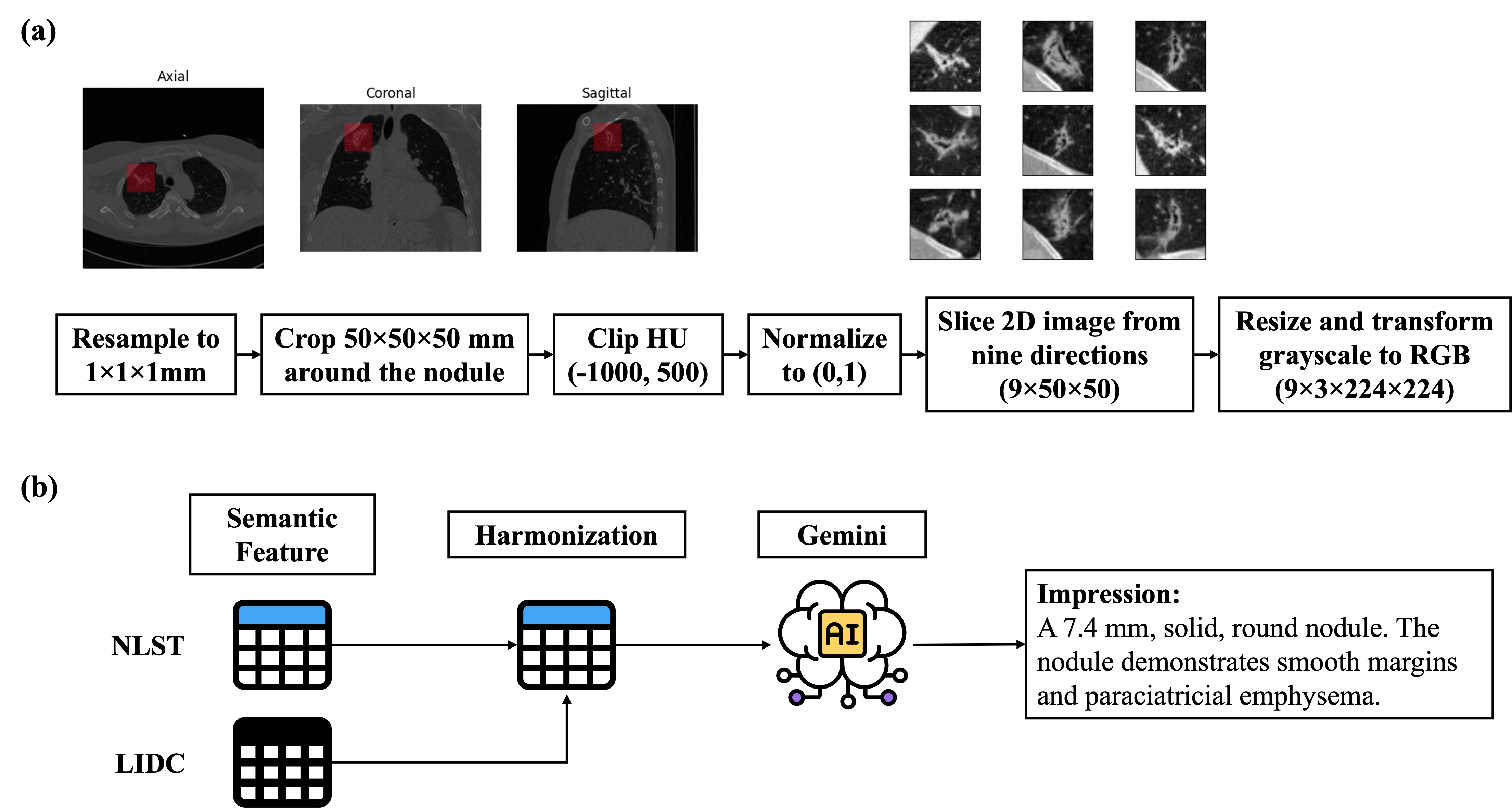}}
\caption{\textbf{CT and Semantic Features Preprocessing Steps.}} (a) For the CT scan, we resampled it to a spacing of $1\times1\times1$ mm and cropped it with a bounding box of size 50 mm. The Hounsfield Unit (HU) values were clipped between $-1000$ and 500, then normalized to a scale of 0 to 1. We sliced the nodule from nine different angles to accommodate the CLIP architecture while preserving most of the nodule's structures. Each of the 2D images was then resized and transformed from grayscale to RGB format to meet the requirement of the CLIP input. (b) For semantic features, we first harmonized LIDC and NLST semantic features, and then input the features into Gemini with a prompt designed to generate text similar to a radiology report.\label{fig1}
\end{figure}

\subsection{Data Preprocessing}
\subsubsection{Semantic feature preprocessing}
\label{semanticfeaturepreprocessing}

There are several challenges when directly aligning tabular semantic and imaging features using the CLIP framework. First, we have limited training data compared to the quantity usually used in contrastive learning, so it is beneficial to use the pretrained CLIP model, which already exhibits a certain degree of alignment between image and text features. Second, tabular data contains numerous missing values. For example, radiologists at UCLA overlooked certain semantic features during annotation. Approximately 71.8\% of annotations are missing for the ``airway cutoff'' feature, while most other features have 0.1\% to 2.0\% missing data (Table \ref{tab:semantic}). Furthermore, the NLST and LIDC datasets contain distinct sets of semantic features, resulting in the missingness of a majority of these features during the harmonization process. Therefore, we proposed a workaround to convert semantic features into text to resolve the challenge above. First, we consulted with radiologists to standardize the semantic features by matching the terms between the two datasets as closely as possible. The harmonization of semantic features from LIDC and NLST data is shown in Table \ref{tab:appendix_harmonization}. We treated features that appear in NLST but not in LIDC as missing values. Second, we converted semantic features into radiology report-like text using Gemini 1.5 Flash \cite{gemini,gemini1.5}, a large language model developed by Google DeepMind (Fig. \ref{fig1}b). Specifically, we input a list of semantic features with a prompt. The generative model was instructed to produce a report including findings and impression sections, emulating a radiology report. The findings section listed all semantic features, whereas the impression section provided a high-level summary of these features in a few sentences. The absence of features was only mentioned within the findings section. To prevent the generation of irrelevant information, explicit instructions were given to exclude any content beyond the provided data. For missing values, while imputation methods typically add noise and bias into data, the textual format allows skipping the missing values, and the downstream language models can generate structured feature embeddings. The exact prompt utilized and an example report generated are shown in \ref{appendix_gemini}. Furthermore, we implemented natural language preprocessing augmentation to substitute the text with synonyms and randomly crop the sentence using the nlpaug Python package \cite{nlpaug}, allowing for greater variations in text during training. During training, we also randomly selected text from the findings and impressions parts.

\subsubsection{CT images preprocessing}
\label{ctimagepreprocessing}
Fig. \ref{fig1}a illustrates the CT preprocessing pipeline. First, we standardized and resampled 3D CT images to a spatial resolution of $1\times1\times1$ mm and placed a $50\times50\times50$ mm bounding box around the nodule. For nodules located at the boundaries, padding was applied to the cropped image to ensure a consistent input size. We chose the size of 50 mm because our study focuses on pulmonary nodules, which are typically smaller than 30 mm. This size ensures that the cropped image includes the entire nodule and its perinodular region, while still preserving sufficient detail to capture small nodules. Then, we clipped the Hounsfield Unit values to the range of $-1000$ to 500 and normalized them to the range of 0 to 1. Since we aimed to use the CLIP model, we converted the 3D nodule crops into 2D images obtained from nine different planes, all passing through the nodule centroid. This approach is referred to as 2.5D, where multiple 2D images are used to approximate 3D information. This is especially important for characterizing nodules, as different areas of the nodule can exhibit varying attributes. Furthermore, since CT scans are grayscale, we repeated the 2D image three times and stacked them together to mimic three RGB channels. Then, we normalized each 2D image using CLIP preprocessing and resized it to be $224 \times 224$ to maintain consistency with the pretrained model's input requirements.

During training, we applied a random jitter of up to 5 mm to the nodule centroid. Using the TorchIO Python package \cite{torchio}, we performed random flipping (with 50\% probability), affine transformations (with up to 10 degrees of rotation), the addition of Gaussian noise (mean = 0, standard deviation = 0.02), and contrast adjustments by raising voxel intensities to powers between $-0.02$ and 0.02.

\subsection{Experimental Setup}
\subsubsection{Model Architecture and Loss Functions}
\label{modelarchitectureandlossfunctions}
\begin{figure}[!t]
\centerline{\includegraphics[width=\textwidth]{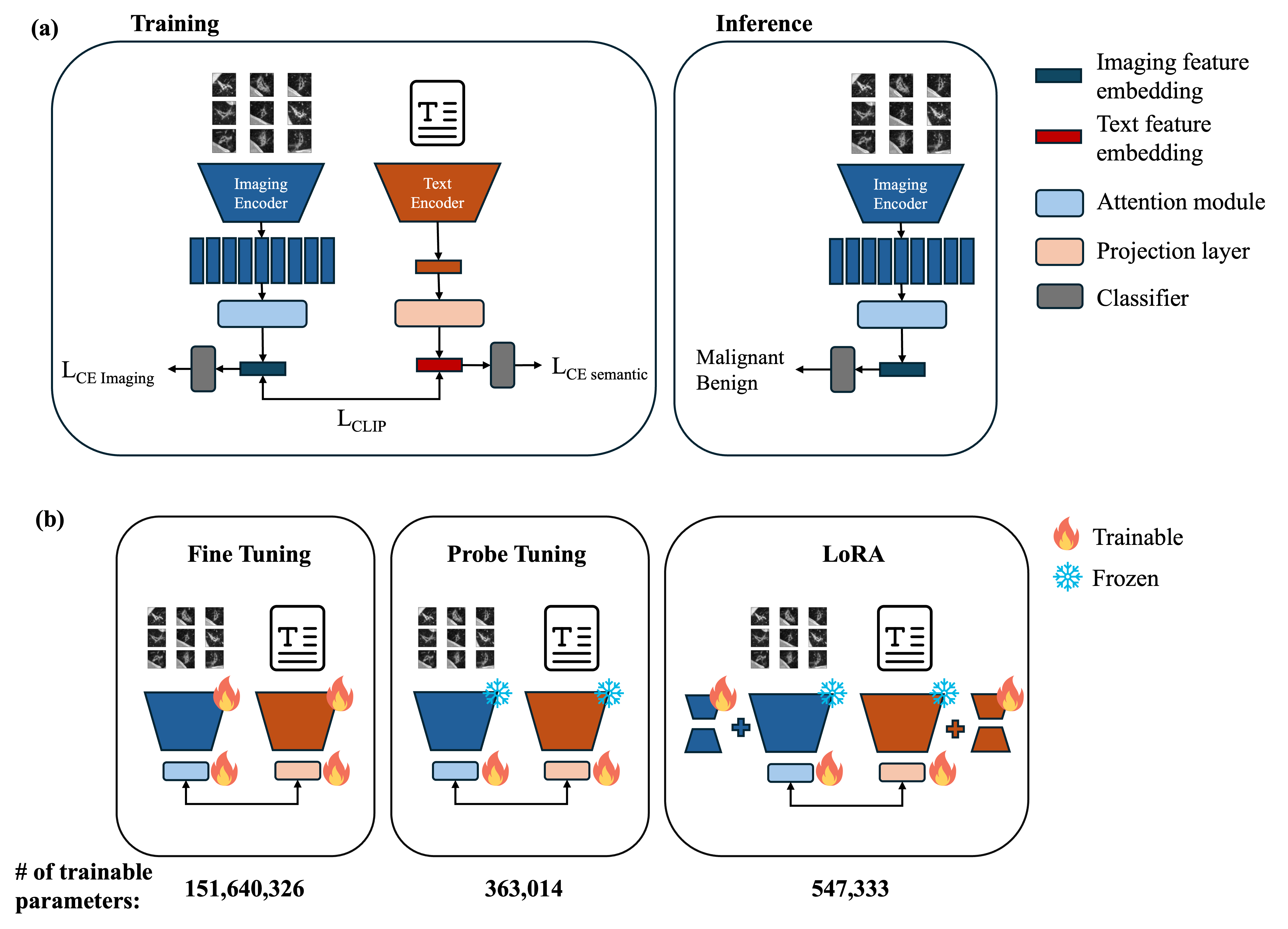}}
\caption{\textbf{CLIP Model Architecture and Fine-tuning Methods.} (a) The pretrained CLIP was fine-tuned to pull the paired imaging and semantic features close together, which allowed the model to learn meaningful relationships between imaging features and semantic features. During training, nodule images sliced in nine different directions from 3D nodule crops were passed into the vision transformer image encoder. The attention-based multiple-instance learning module aggregated the output imaging features to obtain a single embedding. The sentence containing the semantic features was passed into the text encoder to generate the text feature embedding. The visual and text features were then aligned using CLIP loss. Two prediction heads were independently attached after the encoders to predict the one-year lung cancer risk. During inference, only the imaging feature was required, allowing it to be applied universally without the need for a radiologist’s evaluation of the CT scan. (b) We examined the training of the CLIP model using three distinct tuning methods as part of our ablation study. All parameters were fine-tuned (left). Only two projection layers were fine-tuned (middle). A parameter-efficient fine-tuning method, Low-Rank Adaptation (LoRA), which involves inserting trainable low-rank matrices into each layer of the vision and text encoders, was used in our final model (right). While keeping the pretrained weights frozen, we updated the low-rank matrices. We also allowed the projection layers to be fine-tuned.}
\label{fig2}
\end{figure}

The architecture of the model is illustrated in Fig. \ref{fig2}a. The CLIP model was initialized using the weights from OpenAI's CLIP (ViT-B32) and was subsequently fine-tuned to align imaging and semantic features. The 2.5D images were processed through the image encoder, and an attention-based multiple instance learning model was employed to generate the attention scores and aggregate features across nine 2D images \cite{amil}. Simultaneously, we provided Gemini-generated radiology report-like text to the text encoder to obtain feature embeddings. Two projection heads were attached after each encoder to transform the features into 256-dimensional feature embeddings, which were aligned with two InfoNCE losses \cite{infonce}:
\begin{equation}
\mathcal{L}_{\text{InfoNCE-Image}} = 
- \frac{1}{B}\sum_{i=1}^{B} \log\left( \frac{\exp\left( \text{sim}(I_i, S_i)/\tau \right)}{\sum_{j = 1}^{B} \exp\left( \text{sim}(I_i, S_j)/\tau \right)} \right)
\end{equation}

\begin{equation}
\mathcal{L}_{\text{InfoNCE-Semantic}} = 
- \frac{1}{B}\sum_{i=1}^{B} \log\left( \frac{\exp\left( \text{sim}(I_i, S_i)/\tau \right)}{\sum_{k = 1}^{B} \exp\left( \text{sim}(I_k, S_i)/\tau \right)} \right)
\end{equation}

\begin{equation}
\mathcal{L}_{\text{CLIP}} = \frac{1}{2} (\mathcal{L}_{\text{InfoNCE-Image}}+\mathcal{L}_{\text{InfoNCE-Semantic}})
\label{eq:cliploss}
\end{equation}

\noindent where $I_i$ and $S_i$ represent the deep imaging features and the corresponding semantic text features from nodule $i$, which we consider a positive pair. The negative pairs are imaging $I_i$ and all other semantic features in the batches of size B, $S_j$, where $j\neq i$. In addition, negative pairs also include semantic features $S_i$ and all other imaging $I_k$, where $k\neq i$. The cosine similarity (sim) is computed for both positive and negative pairs, and a softmax function is used to transform the similarity into probability, which indicates how likely each pair of imaging and semantic features is a positive match. The temperature, $\tau$, controls the smoothness of the output probability. Rather than being manually tuned, the temperature is set as a learnable parameter and is optimized jointly with the model parameters during training.

This framework allows the image encoder to learn clinically meaningful features by incorporating semantic information, resulting in more robust imaging representations that focus on relevant regions rather than spurious patterns. However, limited training data can hinder effective alignment between modalities. To address this, we introduced two prediction branches, one for imaging features and one for semantic text features, to predict one-year lung cancer diagnoses, encouraging the model to learn diagnosis-relevant features while aligning both modalities. Each branch is trained using cross-entropy loss. Since our data is highly imbalanced, during training, class weighting was applied in the cross-entropy loss functions to assign a higher weight to malignant samples. In addition to the prediction branches, we utilized Low-Rank Adaptation (LoRA), a parameter-efficient technique that resulted in only 0.4\% of the trainable parameters, effectively preventing overfitting \cite{lora}. Specifically, low-rank matrices were incorporated into each query, key, and value layer of the vision and text transformer. The final weights were obtained by summing the pretrained weights with the low-rank matrices. The pretrained weights were kept frozen throughout training, and only the low-rank matrices were updated.

\subsubsection{Model Training and Hyperparameter Settings}
\label{modeltrainingandhyperparametersettings}
We selected 20\% of cases (N = 188) from NLST as the held-out test set. The remaining nodules from NLST were combined with LIDC data for training. We performed 5-fold cross-validation with 80\% and 20\% split on the patient level. We selected the best hyperparameters based on the average Area Under the Receiver Operating Characteristic (AUROC) score across five validation sets.  The best models were trained using three loss functions, including the CLIP loss in Equation \ref{eq:cliploss} and two cross-entropy losses, all with equal weighting. Although the temperature in CLIP loss was learnable during training, we initially set it to 0.03. Optimization was performed using AdamW \cite{adamw} with a learning rate of 0.0001 and a weight decay of 0.1. The model was trained with a batch size of 16. To diversify the nodules within a batch, we selected samples based on the frequency of semantic features, upsampling nodules with rare features. In LoRA, we set the rank of the inserted matrices to 2, the scale factor to 1, and the dropout rate to 0.25.

Since each individual can have multiple nodules, we took the maximum of predicted probabilities to represent the patient-level risk score. We performed Beta calibration \cite{betacal} to prevent the model from being overconfident and to ensure that the probabilities reflect the true likelihoods. Uncalibrated and overconfident probabilities may lead to confusion in clinical interpretations and could potentially result in misleading clinical decisions \cite{guo2017calibration}. Similar to Sybil, we adopted an ensemble approach for inference by averaging the calibrated predictions from all five trained models. 

\subsubsection{Hardware and Computation}
\label{hardwareandcomputation}
Model training and inference were performed on an NVIDIA Quadro RTX 8000 GPU (48 GB memory). Inference averaged 0.011 ± 0.001 seconds per sample, allowing models from five folds to process a nodule in under one second. CPU inference was also efficient, averaging 0.095 ± 0.068 seconds per nodule, using an AMD Ryzen Threadripper 3970X (32 cores, 64 threads) with 256 GB RAM.

\subsection{Evaluation}
\label{evaluation}
We compared our model to three SOTA lung cancer risk prediction models, Sybil, the model by Venkadesh et al., and DeepLungIPN, which were previously introduced in Section \ref{relatedwork}. We evaluated the model's performance in predicting patient-level lung cancer risk within one year using the AUROC and the Area Under the Precision-Recall Curve (AUPRC). While AUROC reflects a model’s overall ability to distinguish between benign and malignant nodules, AUPRC focuses on the precision-recall trade-off and offers a more informative evaluation for imbalanced datasets. This is particularly valuable in cancer screening, where correctly identifying the relatively few malignant cases among many benign ones is critical. In addition to point estimates, we computed 95\% confidence intervals (CIs) by bootstrapping predicted probabilities 10,000 times. Given the importance of high sensitivity in lung cancer screening, we also evaluated model performance at higher recall levels. Since AUROC and AUPRC summarize performance across all thresholds, including those with low recall, we additionally reported false positive rate (FPR) and precision at recall levels closest to 0.6, 0.7, 0.8, and 0.9. This helps assess how well each model minimizes false discoveries while capturing more cancer cases.

Using CLIP, we also obtained predictions on semantic features for the NLST test set using zero-shot inference. For categorical semantic features, we inserted all possible words into the sentence ``This nodule [margin/shape/consistency/...] is [...].'' For binary semantic features, we utilized ``There is [pleural retraction/cyst-like spaces/...]'' to indicate presence and ``No findings'' to suggest absence. We calculated the cosine similarity between the features generated from nodule images and each sentence. The softmax function was applied to the cosine similarity scores. Subsequently, we computed the weighted AUROC score to assess the effectiveness of the zero-shot inference for semantic features with multiple categories and the standard AUROC for binary features.

\subsection{Ablation Studies}
We investigated three different tuning methods to train the CLIP model (Fig. \ref{fig2}b). First, fine-tuning involved optimizing all parameters, but this approach can lead to overfitting. Second, in probe-tuning, we froze the image and text encoders while training only the projection layers. This can decrease the risk of overfitting, but it is less effective when there are significant domain shifts. For instance, we adapted a model originally trained on RGB natural images and captions to work with grayscale medical images and text resembling radiology reports. Third, with the original weights frozen, LoRA, described in Section \ref{modelarchitectureandlossfunctions}, preserves the knowledge from the pretrained model while allowing adaptation to new domains using a small number of trainable parameters.

To compare the model performance with different modalities, we trained a logistic regression model with L1 regularization using the semantic features from the NLST training data and assessed its performance on the NLST test set. We also trained the model using only the CLIP vision encoder on the imaging data and evaluated the outcomes on the NLST test set and three external datasets. 

We calculated the mean and standard deviation of the AUROC and AUPRC scores for each model across five folds. The 95\% CIs were computed using the t-distribution based on the standard error across folds. When comparing only two models, we applied the Wilcoxon signed-rank test at a significance level of 0.05. To evaluate the significance of differences in performance among three models, we conducted statistical testing utilizing the Friedman Chi-Square test. A post-hoc Nemenyi test was used to obtain the pairwise p-value.

\section{Results}
\label{results}

\begin{table}[!ht]
\centering
\small
\caption{\textbf{Model Performance on Lung Cancer Prediction Within One Year.} We conducted a comparative analysis of our model against three SOTA lung cancer risk assessment models, evaluating performance through AUROC and AUPRC scores on a held-out test set, as well as on three external datasets, each encompassing various types of CT scans and patient cohorts. The number of patients is shown for each dataset, with the number of lung cancer cases indicated in brackets. We also reported the 95\% confidence interval (CI) computed by bootstrapping. Metrics are displayed as a point estimate in the first row, followed by [95\% CI lower bound, 95\% CI  upper bound] in the second row in each cell. Sybil and Venkadesh et al. models were not assessed on the NLST test set, as both models were trained on NLST. The performance of Venkadesh et al. on DLCS and UCLA could not be obtained due to issues with data privacy and platform credits. Since DeepLungIPN failed to process some cases across the datasets, the comparison here is not strictly equivalent. In Table \ref{tab:performance_excludemissing}, these cases were excluded, and the corresponding results are reported.}
\resizebox{\textwidth}{!}{
\begin{tabular}{c c c c c c c c c}
\toprule
\multirow{3}{*}{\textbf{Models}} & 
\multicolumn{2}{c}{\textbf{NLST Test Set}} & 
\multicolumn{6}{c}{\textbf{External Datasets}} \\
& \multicolumn{2}{c}{\textbf{\shortstack{NLST \\ N=188 (43)}}} & 
\multicolumn{2}{c}{\textbf{\shortstack{LUNGx \\ 70 (35) }}} & 
\multicolumn{2}{c}{\textbf{\shortstack{DLCS \\ 856 (94)}}} & 
\multicolumn{2}{c}{\textbf{\shortstack{UCLA \\ 51 (28)}}} \\
\cmidrule(lr){2-3} \cmidrule(lr){4-5} \cmidrule(lr){6-7} \cmidrule(lr){8-9}
& \textbf{AUROC} & \textbf{AUPRC} & \textbf{AUROC} & \textbf{AUPRC} & \textbf{AUROC} & \textbf{AUPRC} & \textbf{AUROC} & \textbf{AUPRC} \\
\midrule
\multicolumn{9}{c}{Imaging Only Models} \\
\midrule
Sybil 
& - & - 
& \begin{tabular}{l} 0.662 \\ {[0.525,0.795]} \end{tabular} 
& \begin{tabular}{l} 0.669 \\{[0.502,0.850]} \end{tabular} 
& \begin{tabular}{l} 0.797\\ {[0.742,0.848]} \end{tabular} 
& \begin{tabular}{l}0.468\\{[0.361,0.573]} \end{tabular} 
& \begin{tabular}{l} 0.734 \\ {[0.585,0.865]} \end{tabular}
& \begin{tabular}{l} 0.780 \\{[0.610,0.906]} \end{tabular} \\
Venkadesh et al. & - & - & \begin{tabular}{l} 0.684 \\{[0.554,0.803]} \end{tabular}& \begin{tabular}{l} 0.702 \\{[0.537,0.838]} \end{tabular}& - & - & - & - \\
\midrule
\multicolumn{9}{c}{Imaging + Semantic Features Models} \\
\midrule
DeepLungIPN 
& \begin{tabular}{l} 0.862 \\{[0.794,0.920]} \end{tabular}
& \begin{tabular}{l} 0.679 \\{[0.522,0.815]} \end{tabular}
& \begin{tabular}{l} 0.709 \\{[0.577,0.835]} \end{tabular}
& \begin{tabular}{l} 0.658\\{[0.496,0.856]} \end{tabular}
& \begin{tabular}{l} 0.851\\{[0.803,0.894]} \end{tabular}
& \begin{tabular}{l} \textbf{0.543} \\{[0.435,0.645]} \end{tabular}
& \begin{tabular}{l} 0.624\\{ [0.458,0.779]} \end{tabular}
& \begin{tabular}{l} 0.601 \\{[0.435,0.821]} \end{tabular}\\
Ours (CLIP) 
& \begin{tabular}{l} \textbf{0.901} \\{[0.843,0.950] } \end{tabular}
& \begin{tabular}{l} \textbf{0.776} \\{[0.642,0.880]} \end{tabular}
& \begin{tabular}{l} \textbf{0.771}\\{[0.652,0.880]} \end{tabular}
& \begin{tabular}{l} \textbf{0.813} \\{[0.679,0.914]} \end{tabular}
& \begin{tabular}{l} \textbf{0.861} \\{[0.820,0.897]} \end{tabular}
& \begin{tabular}{l} 0.489\\{[0.382,0.595]} \end{tabular}
& \begin{tabular}{l} \textbf{0.778} \\{[0.643,0.898]} \end{tabular}
& \begin{tabular}{l} \textbf{0.823} \\{[0.661,0.935]} \end{tabular}\\
\bottomrule
\end{tabular}
}
\label{tab:performance}
\end{table}

\subsection{Lung Cancer Risk Prediction}
\label{lungcancerriskprediction}
We present the model performance of one-year lung cancer prediction with AUROC and AUPRC scores in Table \ref{tab:performance}. FPR and precision at recall levels approximating 0.6, 0.7, 0.8, and 0.9 are shown in Table \ref{tab:highrecall}. Since DeepLungIPN failed to generate predictions for some cases, we also report performance metrics excluding those cases to ensure a fairer comparison in \ref{appendix_performance_exclude}.

In the NLST test set, our model surpasses DeepLungIPN (AUROC: 0.862, AUPRC: 0.679), achieving an AUROC of 0.901 (95\% CI: 0.843, 0.950) and an AUPRC of 0.776 (95\% CI: 0.642, 0.880). The CLIP model consistently shows favorable FPR and precision scores across four distinct recall levels.

When evaluated externally on the LUNGx, our model exhibits superior performance with an AUROC of 0.771 (95\% CI: 0.652, 0.880) and an AUPRC of 0.813 (95\% CI: 0.679, 0.914). We also observe lower FPR scores and higher precision scores at a recall level of 0.6, 0.7, and 0.9 for our model. In contrast, DeepLungIPN performs better at a recall of 0.8. The AUROC of our model on the nodule level is 0.769, which outperforms all the reported machine learning-based models in the challenge (mean AUROC of 0.620, with a range of 0.500 to 0.680) \cite{lungx}. The AUC values for the six radiologists range from 0.700 to 0.850 in the observer study. Our model outperforms two of the radiologists in the study but underperforms when compared to the other four.

In the UCLA never-smoker cohort, our model also achieves better performance with an AUROC of 0.778 (95\% CI: 0.643, 0.898) and an AUPRC of 0.823 (95\% CI: 0.661, 0.935). However, DeepLungIPN struggles with the UCLA dataset, with the lowest AUROC of 0.624 (95\% CI: 0.458, 0.779) and AUPRC of 0.601 (95\% CI: 0.435, 0.821), alongside the highest FPR and the lowest precision across all recall levels. Sybil demonstrates fairly strong performance, achieving the highest precision at recall levels of 0.6 and 0.9, while our model consistently presents better FPR and precision at recall levels of 0.7 and 0.8.

In the DLCS, our model achieves the highest AUROC of 0.861 (95\% CI: 0.820, 0.897), although it has a less favorable AUPRC of 0.489 (95\% CI: 0.382, 0.595). DeepLungIPN achieves results comparable to our model, with a lower AUROC of 0.851 (95\% CI: 0.803, 0.894) but a higher AUPRC of 0.543 (95\% CI: 0.435, 0.645). In a more equitable comparison shown in Table \ref{tab:highrecall_excludemissing}, while DeepLungIPN has a lower FPR and higher precision at a recall level of 0.6, our model outperforms it at a recall level of 0.9.

\begin{table}[!t]
\centering
\small
\caption{\textbf{False Positive Rate and Precision at a Given Recall.} High recall is important in lung cancer detection, especially in screening programs, to avoid missing cancer cases. Metrics such as AUROC and AUPRC scores provide an overall evaluation across all thresholds, but may not accurately reflect how the model performs in clinically relevant scenarios. This table presents the false positive rate (FPR) and precision at high recall, ranging from 0.6 to 0.9. Lower FPR and higher precision are optimal. Since DeepLungIPN failed to process some cases across the datasets, the comparison here is not strictly equivalent. In Table \ref{tab:highrecall_excludemissing}, these cases were excluded, and the corresponding results are reported.}
\resizebox{\textwidth}{!}{
\begin{tabular}{l c c c c c c c c}
\toprule
\multirow{2}{*}{\textbf{Models}} & 
\multicolumn{2}{c}{\textbf{Recall $\approx$ 0.6}} &  
\multicolumn{2}{c}{\textbf{Recall $\approx$ 0.7}} &  
\multicolumn{2}{c}{\textbf{Recall $\approx$ 0.8}} & 
\multicolumn{2}{c}{\textbf{Recall $\approx$ 0.9}} \\
\cmidrule(lr){2-3} \cmidrule(lr){4-5} \cmidrule(lr){6-7} \cmidrule(lr){8-9} 
& \textbf{FPR}$\downarrow$ & \textbf{Precision}$\uparrow$ & \textbf{FPR}$\downarrow$ & \textbf{Precision}$\uparrow$ & \textbf{FPR}$\downarrow$ & \textbf{Precision}$\uparrow$ & \textbf{FPR}$\downarrow$ & \textbf{Precision}$\uparrow$ \\
\midrule
\multicolumn{9}{l}{\textbf{NLST Test Set}} \\
\midrule
DeepLungIPN & 0.070 & 0.667 & 0.148 & 0.588 & 0.190 & 0.486 & 0.430 & 0.386 \\
Ours (CLIP) & \textbf{0.034} & \textbf{0.812} & \textbf{0.090} & \textbf{0.638} & \textbf{0.145} & \textbf{0.618} & \textbf{0.338} & \textbf{0.406} \\
\midrule
\multicolumn{9}{l}{\textbf{LUNGx}} \\
\midrule
Sybil & 0.229 & 0.724 & 0.543 & 0.568 & 0.743 & 0.519 & 0.914 & 0.500 \\
Venkadesh et al.  & 0.371 & 0.618 & 0.571 & 0.568 & 0.600 & 0.571 & 0.771 & \textbf{0.542} \\
DeepLungIPN &  0.206 & 0.724 & 0.265 & 0.706 & \textbf{0.471} & \textbf{0.596} & 0.882 & 0.492 \\
Ours (CLIP)& \textbf{0.086} & \textbf{0.875} & \textbf{0.171} & \textbf{0.774} & 0.486 & 0.560 & \textbf{0.743} & \textbf{0.542} \\
\midrule
\multicolumn{9}{l}{\textbf{DLCS}} \\
\midrule

Sybil & 0.136 & 0.354 & 0.227 & 0.273 & 0.444 & 0.187 & 0.580 & 0.153 \\
DeepLungIPN &\textbf{0.097} & \textbf{0.424} & \textbf{0.145} & \textbf{0.361} & \textbf{0.224} & \textbf{0.301} & 0.477 & 0.193 \\
Ours (CLIP) & 0.110 & 0.400 & 0.154 & 0.353 & 0.234 & 0.295 & \textbf{0.379} & \textbf{0.203} \\

\midrule
\multicolumn{9}{l}{\textbf{UCLA}} \\
\midrule
Sybil & \textbf{0.217} & \textbf{0.773} & 0.391 & 0.690 & 0.609 & 0.622 & 0.652 & \textbf{0.634} \\
DeepLungIPN & 0.304 & 0.586 & 0.522 & 0.588 & 0.609 & 0.611 & 0.696 & 0.581 \\
Ours (CLIP) & \textbf{0.217} & 0.739 & \textbf{0.261} & \textbf{0.769} & \textbf{0.261} & \textbf{0.733} & \textbf{0.565} & 0.568 \\

\bottomrule
\end{tabular}
}
\label{tab:highrecall}
\end{table}

\subsection{Model Explainability}
We employed the inherent feature of CLIP, zero-shot inference, to predict semantic features. Table \ref{tab:zeroshot} presents the performance metrics for general, internal, and external features. Our model demonstrates robust AUROC scores for general nodule features, including nodule margin (0.807), margin conspicuity (0.859), and consistency (0.812). Performance in predicting external features is relatively strong, with an AUROC of 0.747 for vascular convergence, 0.840 for pleural attachment, and 0.756 for paracicatrial emphysema. However, the model struggles to predict most internal features, achieving notable results only for cyst-like spaces (0.731) and eccentric calcification (0.794).
\begin{table}[!t]
\centering
\small
\caption{\textbf{Semantic Features Prediction Through Zero-shot Inference.} We presented a subset of the semantic features and the corresponding percentage distribution of each class within these features in the NLST test dataset. Multiple margin characteristics may be annotated for a single nodule. We assessed the performance of predicting semantic features by using weighted AUROC for cases with multiple classes and standard AUROC for those with binary elements. While the prediction shows solid performance in both general and external features, the internal features fail to provide satisfactory zero-shot inference results. This can be attributed to two main factors. First, internal features show a significant imbalance in their presence and absence. Second, the alignment between two modalities may be dominated by semantic features indicating malignancy.} 
\resizebox{\textwidth}{!}{
\begin{tabular}{p{5cm}p{10.5cm}p{1.5cm}}
\toprule
\textbf{Semantic Features} & \textbf{Classes (\%)} & \textbf{AUROC} \\
\midrule
\multicolumn{3}{l}{\textbf{General Features}} \\ 
\midrule
Nodule Margin & Smooth (66.9); Lobulated (22.3); Spiculated (24.3); Ill-defined (19.1); Notched (1.2); N/A (0.8) &0.807 \\
Nodule Consistency &Peri-cystic (2.0); Solid (72.5); Pure ground glass (9.5); Semiconsolidation (8.0); Part-solid (8.0) &0.812 \\
Nodule Shape & Irregular (31.5); Ovoid (35.5); Polygonal (13.5); Round (18.7); N/A (0.8) &0.670 \\
Nodule Margin Conspicuity &  Well marginated (82.9); Poorly marginated (15.9); N/A (1.2)&0.859 \\

\midrule
\multicolumn{3}{l}{\textbf{Internal Features}} \\
\midrule

Nodule Reticulation &  Present (87.6); Absent (12.4) &0.411 \\
Cyst-like Spaces &  Present (13.5); Absent (86.1); N/A (0.4) &0.731 \\
Necrosis & Present (0.4); Absent (99.6) &0.112 \\
Eccentric Calcification&  Present (3.2); Absent (96.4); N/A (0.4) &0.794 \\
Cavitation &  Present (0.4); Absent (99.6) &0.620 \\
Intra-nodular Bronchiectasis&  Present (2.8); Absent (97.2) &0.425 \\
Airway Cutoff&  Present (1.6); Absent (26.7); N/A (71.7) &0.387 \\

\midrule
\multicolumn{3}{l}{\textbf{External Features}} \\
\midrule

Vascular Convergence &  Present (11.2); Absent (88.4); N/A (0.4) &0.747 \\
Pleural Retraction &  Present (20.3) - Mild and Obvious Dimpling; Absent (79.7) &0.689 \\
Pleural Attachment & Present (52.6); Absent (47.4) &0.840 \\
Paracicatricial Emphysema &  Present (12.0); Absent (87.6); N/A (0.4) &0.756 \\
Septal Stretching &  Present (68.9); Absent (30.3); N/A (0.8) &0.670 \\
\bottomrule
\end{tabular}
}
\label{tab:zeroshot}
\end{table}

\subsection{Ablation Studies}

In Table \ref{tab:ablation}, we present the mean and standard deviation of AUROC and AUPRC across five folds for models trained with various tuning methods and modalities. The CLIP model trained with LoRA performs better in predicting lung cancer risk than one trained with fine-tuning and probe-tuning. Furthermore, in the NLST test set, with only the semantic features, the model achieves a mean AUROC of 0.888 (95\% CI: 0.886, 0.890), which is comparable to our model, and a mean AUPRC of 0.780 (95\% CI: 0.772, 0.795), which is the highest among the three trained models. The CLIP model, trained with both semantic and imaging features, outperforms the CLIP vision encoder trained with imaging data only across all datasets.

\begin{table}[!t]
\centering
\small
\caption{\textbf{Performance of Ablation Studies.}} The first three rows display the outcome of the CLIP model, which was trained using various tuning methods. The last three rows display the performance of models trained on different modalities. In each cell, the mean and standard deviation of AUROC scores across five-fold models were presented in the first row. In the second row, the 95\% confidence intervals were computed using the t-distribution based on the standard error across folds. The * superscript denotes a statistically significant difference in performance compared to the best-performing model. P-values are shown in Table \ref{tab:ablation_pval}. Abbreviation: FT --- Fine-Tuning; PT --- Probe-Tuning; LoRA --- Low-Rank Adaptation; S --- Logistic regression model trained on semantic features; V --- CLIP vision encoder trained with imaging only; CLIP --- Contrastive Language–Image Pre-training.
\resizebox{\textwidth}{!}{
\begin{tabular}{c c c c c c c c c}
\toprule
\multirow{3}{*}{} & 
\multicolumn{2}{c}{\textbf{NLST Test Set}} & 
\multicolumn{6}{c}{\textbf{External Datasets}} \\
& \multicolumn{2}{c}{\textbf{\shortstack{NLST \\ N=188 (43)}}} & 
\multicolumn{2}{c}{\textbf{\shortstack{LUNGx \\ 70 (35) }}} & 
\multicolumn{2}{c}{\textbf{\shortstack{DLCS \\ 856 (94)}}} & 
\multicolumn{2}{c}{\textbf{\shortstack{UCLA \\ N = 51 (28)}}} \\
\cmidrule(lr){2-3} \cmidrule(lr){4-5} \cmidrule(lr){6-7} \cmidrule(lr){8-9}
& \textbf{AUROC} & \textbf{AUPRC} & \textbf{AUROC} & \textbf{AUPRC} & \textbf{AUROC} & \textbf{AUPRC} & \textbf{AUROC} & \textbf{AUPRC} \\
\midrule

\multicolumn{9}{c}{\textbf{Training With Different Tuning Methods}} \\
\midrule

FT 
& \begin{tabular}{l} $0.809\pm0.012^*$ \\{[0.793,0.825]} \end{tabular}
& \begin{tabular}{l} $0.533\pm0.050^*$ \\{[0.463,0.603]} \end{tabular} 
& \begin{tabular}{l} $0.656\pm0.040^*$ \\{[0.600,0.712]} \end{tabular}
& \begin{tabular}{l} $0.638\pm0.060^*$ \\{[0.555,0.722]} \end{tabular} 
& \begin{tabular}{l} $0.798\pm0.014^*$  \\{[0.778,0.818]} \end{tabular} 
& \begin{tabular}{l} $0.328\pm0.018^*$  \\{[0.302,0.353]} \end{tabular} 
& \begin{tabular}{l} $0.576\pm0.038^*$ \\{[0.523,0.629]} \end{tabular} 
& \begin{tabular}{l} $0.602\pm0.052^*$ \\{[0.529,0.675]} \end{tabular} \\
PT
& \begin{tabular}{l} $0.839\pm0.018$ \\{[0.814,0.863]} \end{tabular}  
& \begin{tabular}{l} $0.634\pm0.026$ \\{[0.599,0.670]} \end{tabular}   
& \begin{tabular}{l} $0.733\pm0.016$ \\{[0.710,0.755]} \end{tabular}  
& \begin{tabular}{l} $0.728\pm0.031$ \\{[0.684,0.771]} \end{tabular}  
& \begin{tabular}{l} $0.836\pm0.002$ \\{[0.832,0.839]} \end{tabular}  
& \begin{tabular}{l} $0.423\pm0.009$ \\{[0.410,0.436]} \end{tabular}  
& \begin{tabular}{l} $0.722\pm0.017$ \\{[0.698,0.746]} \end{tabular}  
& \begin{tabular}{l} $0.753\pm0.007$ \\{[0.743,0.762]} \end{tabular}  \\
LoRA  
& \begin{tabular}{l} $\textbf{0.889}\pm 0.009$ \\{[0.876,0.902]} \end{tabular}  
& \begin{tabular}{l} $\textbf{0.757}\pm0.017$ \\{[0.733,0.781]} \end{tabular}  
& \begin{tabular}{l} $\textbf{0.763}\pm0.013$ \\{[0.745,0.782]} \end{tabular}  
& \begin{tabular}{l} $\textbf{0.801}\pm0.017$ \\{[0.778,0.825]} \end{tabular}  
& \begin{tabular}{l} $\textbf{0.852}\pm0.010$ \\{[0.838,0.866]} \end{tabular}  
& \begin{tabular}{l} $\textbf{0.461}\pm0.041$ \\{[0.404,0.517]} \end{tabular}  
& \begin{tabular}{l} $\textbf{0.760}\pm0.041$ \\{[0.702,0.817]} \end{tabular}  
& \begin{tabular}{l} $\textbf{0.805}\pm0.035$ \\{[0.756,0.854]} \end{tabular}  \\
\midrule
\multicolumn{9}{c}{\textbf{Training With Different Modalities}} \\
\midrule

S 
& \begin{tabular}{l} $0.888\pm0.002$ \\{[0.886,0.890]} \end{tabular}  
& \begin{tabular}{l} $\textbf{0.780}\pm0.009$ \\{[0.772,0.795]} \end{tabular} & - & - & - & - & - & - \\
V
& \begin{tabular}{l} $0.812\pm0.032^*$ \\{[0.768,0.857]} \end{tabular}  
& \begin{tabular}{l} $0.587\pm0.046^*$ \\{[0.523,0.651]} \end{tabular}  
& \begin{tabular}{l} $0.685\pm0.036$ \\{[0.635,0.735]} \end{tabular}  
& \begin{tabular}{l} $0.718\pm0.034$ \\{[0.671,0.765]} \end{tabular}  
& \begin{tabular}{l} $0.798\pm0.014$ \\{[0.778,0.817]} \end{tabular}  
& \begin{tabular}{l} $0.385\pm0.036$ \\{[0.335,0.436]} \end{tabular}  
& \begin{tabular}{l} $0.685\pm0.046$ \\{[0.621,0.750]} \end{tabular}
& \begin{tabular}{l} $0.709\pm0.037$ \\{[0.657,0.760]} \end{tabular}  \\
CLIP 
& \begin{tabular}{l} $\textbf{0.889}\pm 0.009$ \\{[0.876,0.902]} \end{tabular}  
& \begin{tabular}{l} $0.757\pm0.017$ \\{[0.733,0.781]} \end{tabular}  
& \begin{tabular}{l} $\textbf{0.763}\pm0.013$ \\{[0.745,0.782]} \end{tabular}  
& \begin{tabular}{l} $\textbf{0.801}\pm0.017$ \\{[0.778,0.825]} \end{tabular}  
& \begin{tabular}{l} $\textbf{0.852}\pm0.010$ \\{[0.838,0.866]} \end{tabular}  
& \begin{tabular}{l} $\textbf{0.461}\pm0.041$ \\{[0.404,0.517]} \end{tabular}  
& \begin{tabular}{l} $\textbf{0.760}\pm0.041$ \\{[0.702,0.817]} \end{tabular}  
& \begin{tabular}{l} $\textbf{0.805}\pm0.035$ \\{[0.756,0.854]} \end{tabular}  \\

\bottomrule
\end{tabular}
}
\label{tab:ablation}
\end{table}

\subsection{Error Analysis}

\begin{figure}[!t]
\centerline{\includegraphics[width=\textwidth]{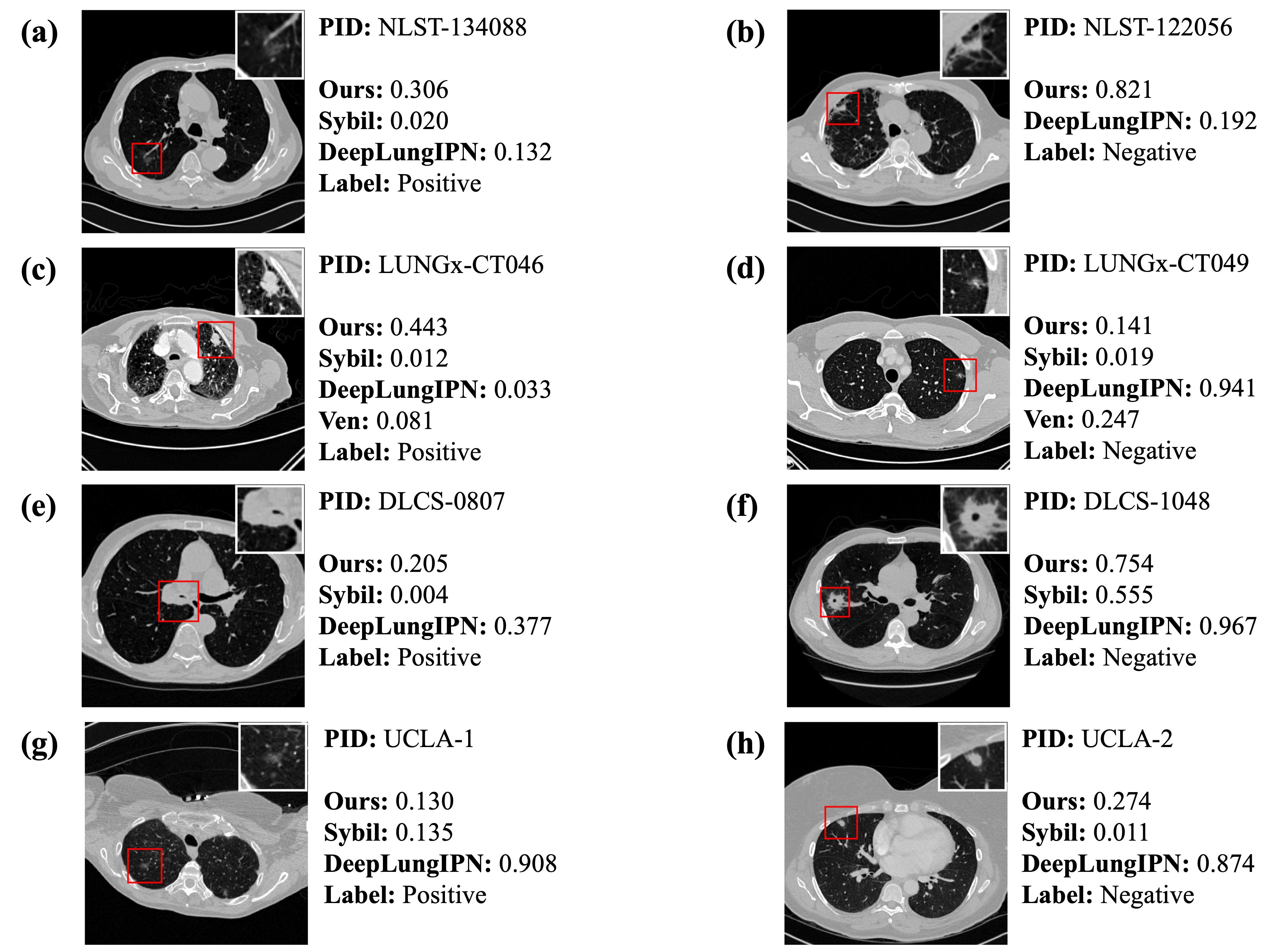}}
\caption{\textbf{Error Analysis.} One lung cancer and one non-lung cancer case from each dataset are presented. We present the CT scan slice corresponding to the middle of the nodule, highlighted with a red bounding box. In the top right corner, we place the magnified view of the nodule for clearer visualization. Certain features may not be fully appreciable in the single slice but are visible when viewing the whole series. Patient-level risk score from each model is shown beside each image. The scale and range of the predicted probability across different models can vary significantly. The probabilities from Sybil, Venkadesh et al., and our model have been calibrated. Therefore, the predicted probability from these three models matches the observed frequency. Sybil risk scores for NLST are listed only if the case is included in Sybil's test data.}
\label{fig3}
\end{figure}

In Fig. \ref{fig3}, we present a CT scan from a lung cancer patient and a non-lung cancer patient for each dataset. More examples are shown in Fig. \ref{fig3_appendix}. It is important to note that both Sybil and DeepLungIPN use only the CT scan as input, whereas our model and that of Venkadesh et al. incorporate both the CT scan and the nodule location as inputs. Although the pipeline in DeepLungIPN involves a nodule detection step, the detected nodules may not correspond to the annotated nodules shown in the figure.

In Fig. \ref{fig3}a and c, all models fail to identify the patient as high-risk even though they are diagnosed with lung cancer within a year. In Fig. \ref{fig3}a, the nodule crop shows a poorly marginated, pure ground glass nodule with an irregular margin. The malignant nodule in Fig. \ref{fig3}c is spiculated with a complex shape, septal stretching, and vascular convergence. Our model outputs relatively higher risk scores of 0.306 and 0.443, in contrast to the extremely low scores given by other models. Fig. \ref{fig3}e illustrates a challenging and uncommon case where the nodule is within the right main stem bronchus. All models assign a relatively lower score while the nodule is malignant. In Fig. \ref{fig3}g, a poorly marginated pure ground glass nodule is presented. Among the models, only DeepLungIPN successfully assigns a high risk score to this malignant case, while the others fail.

Fig. \ref{fig3}b shows a non-cancerous case with a lobulated and serrated nodule exhibiting complex shape and pleural retraction. While DeepLungIPN correctly identifies the case as low risk, our model assigns a high risk score of 0.821. In contrast, Fig. \ref{fig3}d and h illustrate two negative cases where DeepLungIPN assigns unusually high risk scores, whereas our model and others provide more conservative estimates. This discrepancy may stem from DeepLungIPN’s miscalibration, leading to overconfident predictions. Additionally, visual inspection revealed that DeepLungIPN sometimes detects non-nodule regions, which may contribute to inflated risk scores. In Fig. \ref{fig3}f, although the biopsy confirmed a benign finding, the nodule appears highly suspicious due to its large size, irregular shape, and spiculated margins. All models, including ours, assign high risk scores to this case. Radiologists also classified it as Lung Imaging Reporting and Data System (Lung-RADS) category 4X, indicating a high suspicion of malignancy with additional concerning imaging features \cite{lungrads}.

\section{Discussion}
\label{discussion}
In this study, we trained a CLIP model to align CT imaging features with semantic features to facilitate better lung cancer prediction. We preprocessed both image and semantic features to ensure alignment with the pretrained CLIP format. To tackle the challenge of fine-tuning with a small dataset, we integrated LoRA and supervised branches into CLIP. Our model outperforms the SOTA models, showing consistent robustness across external datasets.

In the test and external datasets, we included a variety of CT scans collected in different scenarios. The performance of Sybil and Venkadesh et al. noticeably decreases when applied to the LUNGx dataset, which includes diagnostic CT and CECT scans. This decrease can be attributed to the fact that both models were trained on the NLST dataset, which only contains LDCT, and the resulting distribution shift adversely impacted their performance. However, our model remains robust even though we did not explicitly train on CECT. Additionally, our model exhibits superior performance on the UCLA dataset, which comprises incidental nodules from non-smoking patients. Nevertheless, the sample size in the external dataset may be limited, and the range of the 95\% CI is quite large. Our model achieves the highest performance in terms of AUROC on the DLCS dataset, but yields a lower overall AUPRC. One potential cause of suboptimal results of our model in DLCS is that, although radiologists review the detected nodules from the algorithm, some still do not correspond to true nodules. As our model was trained exclusively on radiologists-verified nodules, its performance can be unstable when evaluated on regions that do not contain confirmed nodules.

There are several benefits of incorporating semantic features into training. First, the use of semantic features during training guides imaging-based models to learn clinically meaningful patterns of nodules, rather than relying on spurious correlations. In a previous study \cite{generalistmodel}, we investigated the characteristics and robustness of various imaging features by selecting prompts based on feature similarity for nodule segmentation in generalist foundation models. We extracted imaging features from our CLIP model and compared them against pixel intensity, radiomics features, and imaging biomarker foundation features for prompt selection. The results demonstrated that CLIP imaging features are more effective in focusing on relevant nodule characteristics, which led to superior segmentation accuracy and robustness on external datasets using generalist segmentation models. Second, even though the model trained exclusively on semantic features yields comparable or surpassing results to CLIP, using semantic features for prediction is not suitable for clinical use due to subjectivity and variability in semantic feature annotations among radiologists, and the added burden of requiring detailed semantic assessments from radiologists. On the other hand, even though the semantic features were utilized during training in CLIP, our CLIP model only requires CT images as input data at the inference stage, with potential applications across more diverse clinical settings. Third, incorporating domain knowledge about nodule characteristics makes the training process more data-efficient. Our model achieves performance comparable to or surpassing imaging-based SOTA models despite being trained on limited data. Our model utilized approximately 2,000 cases, whereas Sybil and Venkadesh et al. were trained on approximately 15,000 and 10,000 cases, respectively.

The zero-shot inference capability of the CLIP model improves model explainability, allowing our model to predict semantic features without requiring explicit training. This capability can help clinicians understand the underlying meaning of model predictions. Our model achieves high performance in predicting various general nodule characteristics and external features, but predictions for other features, like nodule shape and most internal features, did not perform well. First, as shown in Table \ref{tab:semantic}, most of the internal features exhibit a significant imbalance in the presence and absence cases. For example, only 0.4\% of the nodules present with necrosis, and 2.5\% of nodules have an airway cutoff. Despite upsampling cases with rare features, the limited number of such instances may result in underrepresentation for certain subgroups and limit the model's effectiveness in differentiating such features. Second, not all features contribute equally to predicting nodule malignancy. It is possible that general and external features are more informative than internal features. In addition, a well-known limitation of CLIP, its lack of fine-grained alignment, could cause the variability in zero-shot semantic feature prediction performance. Imaging and semantic features are converted into a single embedding and subsequently aligned. Therefore, some semantic features can overshadow others during the training process. In the future, our goal is to develop and validate the model further to facilitate more granular alignment.  

While we used Gemini to convert tabular semantic features into radiology report–like text, we did not validate the generated outputs against actual radiology reports. However, our primary goal is not to replicate exact clinical language, but to show the promise of using the CLIP model to effectively align text describing nodule characteristics with nodule images for improved lung cancer diagnosis. Future work could incorporate millions of institution-specific radiology reports, containing detailed descriptions of nodule characteristics, to further enhance the training of CLIP-based foundation models. In addition, while we extracted nodules from nine distinct directions to preserve 3D characteristics to the greatest extent possible, certain features could still be overlooked during this process. Moreover, since our model focused exclusively on the nodule region, we did not include broader lung features like fibrosis and emphysema, which could also serve as risk factors for lung cancer. Furthermore, we did not evaluate the model developed by Venkadesh et al. on the DLCS and UCLA datasets due to limitations of running the model and sharing local data in a public cloud environment.

\section{Conclusion}
We presented the benefits of integrating semantic features using the CLIP model, which allows the imaging encoder to learn more clinically significant and robust features. This model has achieved strong performance in predicting lung cancer risk across external datasets with various patient cohorts and CT scan types. Additionally, our model provides explainability regarding nodule characteristics through zero-shot inference, equipping clinicians with better insights into how the model makes its predictions. Although it was trained with semantic features, our model does not require manual annotation from radiologists, which enhances its scalability in diverse clinical settings.

\section*{CRediT authorship contribution statement}

\noindent\textbf{Luoting Zhuang:} Writing - Original Draft, Writing - Review \& Editing, Data Curation, Software, Methodology, Formal analysis, Visualization, Validation, Investigation, Conceptualization. \textbf{Seyed Mohammad Hossein Tabatabaei:} Data Curation, Writing – review \& editing. \textbf{Ramin Salehi-Rad:} Data Curation, Writing – review \& editing. \textbf{Linh M. Tran:} Data Curation, Writing – review \& editing. \textbf{Denise Aberle:} Data Curation, Writing – review \& editing, Funding acquisition, Conceptualization. \textbf{Ashley Prosper:} Data Curation, Writing – review \& editing, Conceptualization. \textbf{William Hsu:} Data Curation, Writing – review \& editing, Supervision, Project administration, Investigation, Funding acquisition, Conceptualization, Resources.

\section*{Declaration of competing interest}
\noindent William Hsu reports funding support from the National Institutes of Health, Agency for Healthcare Research and Quality, Early Diagnostics Inc, personal fees from the Radiological Society of North America related to editorial board work, and consulting fees from LungLife AI, Inc. Ashley E. Prosper reports funding support from the National Institutes of Health. Linh M. Tran reports funding support from the Department of Veterans Affairs Merit Review. Denise R. Aberle reports funding support from the National Institutes of Health. If there are other authors, they declare that they have no known competing financial interests or personal relationships that could have appeared to influence the work reported in this paper. 

\section*{Code Availability}
\noindent The code and model weights are available at \url{https://github.com/luotingzhuang/CLIP_nodule}.

\section*{Acknowledgments}
\noindent We gratefully acknowledge the support of NIH/National Cancer Institute U2C CA271898 (to R.S-R., L.T., D.A., A.P., W.H.), U01 CA233370 (to L.Z., D.A., A.P., W.H.), the V Foundation (to D.A., W.H.), and the Department of Veterans Affairs Merit Review I01BX005721 (to L.T.). The content is solely the responsibility of the authors and does not necessarily represent the official views of the National Institutes of Health. \\

\noindent The authors acknowledge the National Cancer Institute and the Foundation for the National Institutes of Health, and their critical role in the creation of the free publicly available LIDC/IDRI Database used in this study. LUNGx data used in this research were obtained from The Cancer Imaging Archive (TCIA) sponsored by the SPIE, NCI/NIH, AAPM, and The University of Chicago.

\setcounter{page}{1}

\appendix
\renewcommand{\thetable}{A\arabic{table}}
\renewcommand{\thefigure}{A\arabic{figure}}

\setcounter{table}{0} 
\setcounter{section}{0}
\setcounter{figure}{0}

\clearpage
\section{Supplementary data}

\subsection{Semantic Features Annotated}
\begin{table}[H]
\centering
\small
\caption{\textbf{Semantic Features Annotated by Radiologists for 1,261 Nodules from NLST Data.} We reported all semantic features and the percentage distribution of each class within these features for the entire NLST dataset annotated by radiologists at UCLA Health. Multiple margin characteristics may be annotated for a single nodule. }
\resizebox{\textwidth}{!}{
\begin{tabular}{p{5cm}p{13cm}}
\toprule
\textbf{Feature} &  \textbf{Classes (\%)} \\
\midrule
\multicolumn{2}{l}{\textbf{General Features}} \\ 
\midrule
Longest Axial Diameter  & 3.8-10mm (60.6); 10-20 (31.2); 20-30 (7.1); 30-40 (0.7); 40-50 (0.2); 50-59.3 (0.2) \\
Short Diameter  & 2.0-10mm (79.4); 10-20 (18.4); 20-30 (1.6); 30-40 (0.5); 40-48.8 (0.1) \\
Nodule Margin & Smooth (65.9); Lobulated (23.2); Spiculated (24.7); Ill-defined (20.6); Notched (0.2); N/A (2.0)  \\
Nodule Consistency &Peri-cystic (1.9); Solid (70.0); Pure ground glass (9.4); Semiconsolidation (8.0); Part-solid (10.6); N/A (0.1) \\
Nodule Shape & Irregular (37.3); Ovoid (31.6); Polygonal (11.3); Round (18.9); N/A (0.9)  \\
Nodule Margin Conspicuity &  Well marginated (80.8); Poorly marginated (18.1); N/A (1.1) \\

\midrule
\multicolumn{2}{l}{\textbf{Internal Features}} \\
\midrule

Nodule Reticulation &  Present (87.1); Absent (12.5); N/A (0.4)  \\
Cyst-like Spaces &  Present (13.8); Absent (85.6); N/A (0.6) \\
Necrosis & Present (0.4); Absent (99.6) \\
Eccentric Calcification&  Present (3.2); Absent (96.4); N/A (0.4) \\
Cavitation &  Present (0.4); Absent (99.6)  \\
Intra-nodular Bronchiectasis&  Present (3.0); Absent (96.8); N/A (0.2) \\
Airway Cutoff&  Present (2.5); Absent (25.7); N/A (71.8)  \\

\midrule
\multicolumn{2}{l}{\textbf{External Features}} \\
\midrule

Vascular Convergence &  Present (12.8); Absent (86.8); N/A (0.4) \\
Pleural Retraction &  Present (21.7) - Mild and Obvious Dimpling; Absent (78.1); N/A (0.2) \\
Pleural Attachment & Present (51.5); Absent (48.2); N/A (0.3) \\
Paracicatricial Emphysema &  Present (11.1); Absent (88.5); N/A (0.4) \\
Septal Stretching &  Present (67.3); Absent (32.5); N/A (0.2) \\

\midrule
\multicolumn{2}{l}{\textbf{General Assessment}} \\
\midrule
Level of Suspicion of Lung Cancer  & Very Low (30.1); Moderately Low (22.6); Intermediate (20.3); Moderately High (12.8); High (13.4); N/A (0.8)\\

\bottomrule
\end{tabular}
}
\label{tab:semantic}
\end{table}

\clearpage

\subsection{Text Generation from Tabular Data Using Gemini}
\label{appendix_gemini}
\noindent \textbf{Prompt:} 
\begin{tcolorbox}[colback=gray!5, colframe=black, boxrule=0.5pt, arc=4pt]

Think as if you are a radiologist, you have a table/dictionary for lung nodule evaluation, and you want to transform it into radiology reports. Here is the dictionary \{...\}. Just show me the report. Do not add any additional information. In English! You should have an impression and findings part. Findings are bullet points listing each of the features separately. These features should be in random order. Combine nodule margins and additional nodule margins. If there is a missing value, just ignore it. For impression, do a quick summary of the findings in sentence and then state the suspiciousness of lung cancer. You can mention the absences in the findings but do not mention those in the impression.
\end{tcolorbox}
\clearpage
\noindent \textbf{Example: }

\begin{tcolorbox}[colback=gray!5, colframe=black, boxrule=0.5pt, arc=4pt]
Findings:\\
- Axial location: Central\\
- Longest axial diameter (mm): 17.0\\
- Short diameter (mm): 5.1\\
- Nodule margins: Ill-defined, Spiculated, Serrated\\
- Nodule shape: Complex, Irregular\\
- Nodule consistency: Part-solid\\
- Nodule reticulation: Present\\
- Cyst-like spaces: Absent\\
- Intra nodular bronchiectasis: Absent\\
- Necrosis: Absent\\
- Cavitation: Absent\\
- Eccentric calcification: Absent\\
- Airway cutoff: Absent\\
- Pleural attachment: Present\\
- Pleural retraction: Mild dimpling\\
- Vascular convergence: Present\\
- Septal stretching: Present\\
- Paracicatricial emphysema: Absent\\
\\
Impression:\\
A 17.0 $\times$ 5.1 mm, central, part-solid nodule is identified. The nodule demonstrates complex, spiculated margins and is associated with pleural attachment, mild dimpling, vascular convergence, and septal stretching. The level of suspicion for lung cancer is moderately high. 
\end{tcolorbox}

\clearpage

\subsection{LIDC and NLST Semantic Features Harmonization}
\label{appendix_harmonization}

\begin{table}[H]
\centering
\caption{\textbf{Harmonization of LIDC and NLST semantic features.} We consulted radiologists at our institution to map the semantic features in LIDC to the corresponding semantic features in our internal NLST database. Semantic features present in our in-house data but absent in LIDC are considered missing and were excluded during text generation.}
\resizebox{\textwidth}{!}{

\begin{tabular}{lll}
\toprule
\textbf{LIDC Semantic Features} & \textbf{Radiologists' Entry} & \textbf{Mapped to NLST Semantic Features} \\
\midrule
\multirow[t]{2}{*}{Internal Structure} & ``Air'' & Cyst-like spaces = ``Present'' \\
                         & All Others & Cyst-like spaces = ``Absent'' \\
\multirow[t]{2}{*}{Calcification} & ``Non central appearance'' & Eccentric Calcification = ``Present'' \\
                         & All Others & Eccentric Calcification = ``Absent'' \\
\multirow[t]{2}{*}{Sphericity} & $>3$ & Nodule Shape = ``Round'' \\
                         & $\leq3$ & Nodule Shape = ``Ovoid'' \\
\multirow[t]{2}{*}{Margin} & $\geq3$ & Nodule Margin Conspicuity = ``Well marginated'' \\
                         & $<3$ & Nodule Margin Conspicuity = ``Poorly marginated'' \\
\multirow[t]{1}{*}{Lobulation} & $\geq3$ & Nodule Margins = ``Lobulated''\\
\multirow[t]{1}{*}{Spiculation} & $\geq3$ & Nodule Margins = ``Spiculated''\\
\multirow[t]{2}{*}{Texture} & $>4$ & Nodule Consistency = ``Solid''\\
                         & $=2,3,4$ & Nodule Consistency = ``Part-solid'' \\
                         & $<2$ & Nodule Consistency = ``Pure ground glass'' \\
                         \bottomrule
\end{tabular}
}
\label{tab:appendix_harmonization}
\end{table}

\clearpage

\subsection{Model Performance Removing Cases Failed in DeepLungIPN}
To ensure a fair comparison with DeepLungIPN, which failed to predict lung cancer risk for certain cases, we removed the failed cases and reported AUROC and AUPRC scores in Table \ref{tab:performance_excludemissing} and FPR and Precision at different recall levels in Table \ref{tab:highrecall_excludemissing}.
\label{appendix_performance_exclude}

\begin{table}[H]
\centering
\small
\caption{\textbf{Model Performance on Lung Cancer Prediction Within One Year.}} 
\resizebox{\textwidth}{!}{
\begin{tabular}{c c c c c c c c c}
\toprule
\multirow{3}{*}{\textbf{Models}} & 
\multicolumn{2}{c}{\textbf{NLST Test Set}} & 
\multicolumn{6}{c}{\textbf{External Datasets}} \\
& \multicolumn{2}{c}{\textbf{\shortstack{NLST \\ N=186 (43)}}} & 
\multicolumn{2}{c}{\textbf{\shortstack{LUNGx \\ 69 (35) }}} & 
\multicolumn{2}{c}{\textbf{\shortstack{DLCS \\ 838 (94)}}} & 
\multicolumn{2}{c}{\textbf{\shortstack{UCLA \\ 51 (28)}}} \\
\cmidrule(lr){2-3} \cmidrule(lr){4-5} \cmidrule(lr){6-7} \cmidrule(lr){8-9}
& \textbf{AUROC} & \textbf{AUPRC} & \textbf{AUROC} & \textbf{AUPRC} & \textbf{AUROC} & \textbf{AUPRC} & \textbf{AUROC} & \textbf{AUPRC} \\
\midrule
\multicolumn{9}{c}{Imaging Only Models} \\
\midrule
Sybil 
& - & - 
& \begin{tabular}{l} 0.660 \\ {[0.523,0.790]} \end{tabular} 
& \begin{tabular}{l} 0.670 \\{[0.503,0.850]} \end{tabular} 
& \begin{tabular}{l} 0.798\\ {[0.743,0.849]} \end{tabular} 
& \begin{tabular}{l} 0.472\\{[0.365,0.576]} \end{tabular} 
& \begin{tabular}{l} 0.734 \\ {[0.585,0.865]} \end{tabular}
& \begin{tabular}{l} 0.780 \\{[0.610,0.906]} \end{tabular} \\
Venkadesh et al. & - & - & \begin{tabular}{l} 0.675 \\{[0.543,0.800]} \end{tabular}& \begin{tabular}{l} 0.703 \\{[0.538,0.840]} \end{tabular}& - & - & - & - \\
\midrule
\multicolumn{9}{c}{Imaging + Semantic Features Models} \\
\midrule
DeepLungIPN 
& \begin{tabular}{l} 0.862 \\{[0.794,0.920]} \end{tabular}
& \begin{tabular}{l} 0.679 \\{[0.522,0.815]} \end{tabular}
& \begin{tabular}{l} 0.709 \\{[0.577,0.835]} \end{tabular}
& \begin{tabular}{l} 0.658\\{[0.496,0.856]} \end{tabular}
& \begin{tabular}{l} 0.851\\{[0.803,0.894]} \end{tabular}
& \begin{tabular}{l} \textbf{0.543} \\{[0.436,0.645]} \end{tabular}
& \begin{tabular}{l} 0.624\\{ [0.458,0.779]} \end{tabular}
& \begin{tabular}{l} 0.601 \\{[0.435,0.821]} \end{tabular}\\
Ours (CLIP)  
& \begin{tabular}{l} \textbf{0.903} \\{[0.845,0.951] } \end{tabular}
& \begin{tabular}{l} \textbf{0.779} \\{[0.638,0.884]} \end{tabular}
& \begin{tabular}{l} \textbf{0.764}\\{[0.641,0.876]} \end{tabular}
& \begin{tabular}{l} \textbf{0.814} \\{[0.680,0.912]} \end{tabular}
& \begin{tabular}{l} \textbf{0.862} \\{[0.821,0.899]} \end{tabular}
& \begin{tabular}{l} 0.498\\{[0.389,0.604]} \end{tabular}
& \begin{tabular}{l} \textbf{0.778} \\{[0.643,0.898]} \end{tabular}
& \begin{tabular}{l} \textbf{0.823} \\{[0.661,0.935]} \end{tabular}\\
\bottomrule
\end{tabular}
}
\label{tab:performance_excludemissing}
\end{table}

\begin{table}[H]
\centering
\small
\caption{\textbf{False Positive Rate and Precision at a Given Recall.}}
\resizebox{\textwidth}{!}{
\begin{tabular}{l c c c c c c c c}
\toprule
\multirow{2}{*}{\textbf{Models}} & 
\multicolumn{2}{c}{\textbf{Recall $\approx$ 0.6}} &  
\multicolumn{2}{c}{\textbf{Recall $\approx$ 0.7}} &  
\multicolumn{2}{c}{\textbf{Recall $\approx$ 0.8}} & 
\multicolumn{2}{c}{\textbf{Recall $\approx$ 0.9}} \\
\cmidrule(lr){2-3} \cmidrule(lr){4-5} \cmidrule(lr){6-7} \cmidrule(lr){8-9} 
& \textbf{FPR}$\downarrow$ & \textbf{Precision}$\uparrow$ & \textbf{FPR}$\downarrow$ & \textbf{Precision}$\uparrow$ & \textbf{FPR}$\downarrow$ & \textbf{Precision}$\uparrow$ & \textbf{FPR}$\downarrow$ & \textbf{Precision}$\uparrow$ \\
\midrule
\multicolumn{9}{l}{\textbf{NLST Test Set}} \\
\midrule
DeepLungIPN &0.070 & 0.667 & 0.148 & 0.588 & 0.190 & 0.486 & 0.430 & 0.386 \\
Ours (CLIP) & \textbf{0.035} & \textbf{0.812} & \textbf{0.092} & \textbf{0.638} &\textbf{ 0.141} & \textbf{0.630} &\textbf{ 0.331 }&\textbf{ 0.419} \\
\midrule
\multicolumn{9}{l}{\textbf{LUNGx}} \\
\midrule
Sybil & 0.235 & 0.724 & 0.294 & 0.615 & 0.735 & 0.483 & 0.912 & 0.500 \\
Venkadesh et al.  & 0.382 & 0.600 & 0.441 & 0.571 & 0.618 & 0.560 & \textbf{0.676} & \textbf{0.542} \\
DeepLungIPN &  0.206 & 0.724 & 0.265 & 0.706 & \textbf{0.471 }& \textbf{0.596} & 0.882 & 0.492 \\
Ours (CLIP) & \textbf{0.088 }& \textbf{0.875} &\textbf{ 0.176 }& \textbf{0.774} & 0.500 & 0.560 & 0.765 & \textbf{0.542 }\\
\midrule
\multicolumn{9}{l}{\textbf{DLCS}} \\
\midrule
Sybil & 0.132 & 0.368 & 0.223 & 0.281 & 0.438 & 0.193 & 0.578 & 0.157 \\
DeepLungIPN & \textbf{0.097} & \textbf{0.424 }& \textbf{0.145} & 0.361 & \textbf{0.224} & 0.301 & 0.477 & 0.193 \\
Ours (CLIP) & 0.108 & 0.412 & 0.151 & \textbf{0.363} & 0.228 & \textbf{0.305} & \textbf{0.378} & \textbf{0.208} \\
\midrule
\multicolumn{9}{l}{\textbf{UCLA}} \\
\midrule
Sybil &\textbf{0.217} & \textbf{0.773} & 0.391 & 0.690 & 0.609 & 0.622 & 0.652 & \textbf{0.634} \\
DeepLungIPN & 0.304 & 0.586 & 0.522 & 0.588 & 0.609 & 0.611 & 0.696 & 0.581 \\
Ours (CLIP) & \textbf{0.217} & 0.739 & \textbf{0.261} & \textbf{0.769} & \textbf{0.261} & \textbf{0.733} & \textbf{0.565} & 0.568 \\

\bottomrule
\end{tabular}
}
\label{tab:highrecall_excludemissing}
\end{table}

\clearpage
\subsection{Result of Ablation Studies: P-values}
\label{appendix_pval}

\begin{table}[H]
\centering
\small
\caption{\textbf{Hypothesis Testing P-values in Ablation Studies.} This table shows the p-values for hypothesis testing performed in ablation studies. The * superscript denotes a statistically significant difference at a significance level of 0.05. To compare more than three models, we used the Friedman Chi-Square test with a post-hoc Nemenyi test. For comparing two models, we conducted the Wilcoxon signed-rank test. It is important to note that since we only have five samples, one from each fold, the lowest p-value for the Wilcoxon signed-rank test is 0.063, so it will never yield a significant result. Abbreviations: F - Fine-Tuning; P - Probe-Tuning; L - Low-Rank Adaptation; S - Logistic regression model trained on semantic features; V - CLIP vision encoder trained with imaging only; CLIP - Contrastive Language–Image Pre-training.}
\resizebox{\textwidth}{!}{
\begin{tabular}{c c c c c c c c c}
\toprule
\multirow{3}{*}{} & 
\multicolumn{2}{c}{\textbf{NLST Test Set}} & 
\multicolumn{6}{c}{\textbf{External Datasets}} \\
& \multicolumn{2}{c}{\textbf{\shortstack{NLST \\ N=188 (43)}}} & 
\multicolumn{2}{c}{\textbf{\shortstack{LUNGx \\ 70 (35) }}} & 
\multicolumn{2}{c}{\textbf{\shortstack{DLCS \\ 856 (94)}}} & 
\multicolumn{2}{c}{\textbf{\shortstack{UCLA \\ 51 (28)}}} \\
\cmidrule(lr){2-3} \cmidrule(lr){4-5} \cmidrule(lr){6-7} \cmidrule(lr){8-9}
& \textbf{AUROC} & \textbf{AUPRC} & \textbf{AUROC} & \textbf{AUPRC} & \textbf{AUROC} & \textbf{AUPRC} & \textbf{AUROC} & \textbf{AUPRC} \\
\midrule
\multicolumn{9}{c}{\textbf{CLIP Training With Different Tuning Methods}} \\
\midrule
F vs P & 0.609 & 0.254 & 0.139 & 0.254& 0.139 & 0.139 & 0.139 & 0.139\\
F vs L & 0.012* & 0.004*  & 0.012* & 0.004*& 0.012* & 0.012* & 0.012* & 0.012*\\
P vs L  & 0.139 & 0.254 & 0.609 & 0.254 & 0.609 & 0.609 & 0.609 & 0.609\\
\midrule
\multicolumn{9}{c}{\textbf{Training With Different Modalities}} \\
\midrule
S vs V & 0.069 & 0.012* & - & - & - & - & - & - \\
S vs CLIP & 0.946 & 0.609& - & - & - & - & - & - \\
V vs CLIP & 0.031* & 0.139& 0.063& 0.063 & 0.063& 0.125 & 0.063 & 0.063 \\
\bottomrule
\end{tabular}
}
\label{tab:ablation_pval}
\end{table}

\clearpage
\subsection{Error Analysis: More Examples}

\label{appendix_erroranalysis}
\begin{figure}[H]
\centerline{\includegraphics[width=\textwidth]{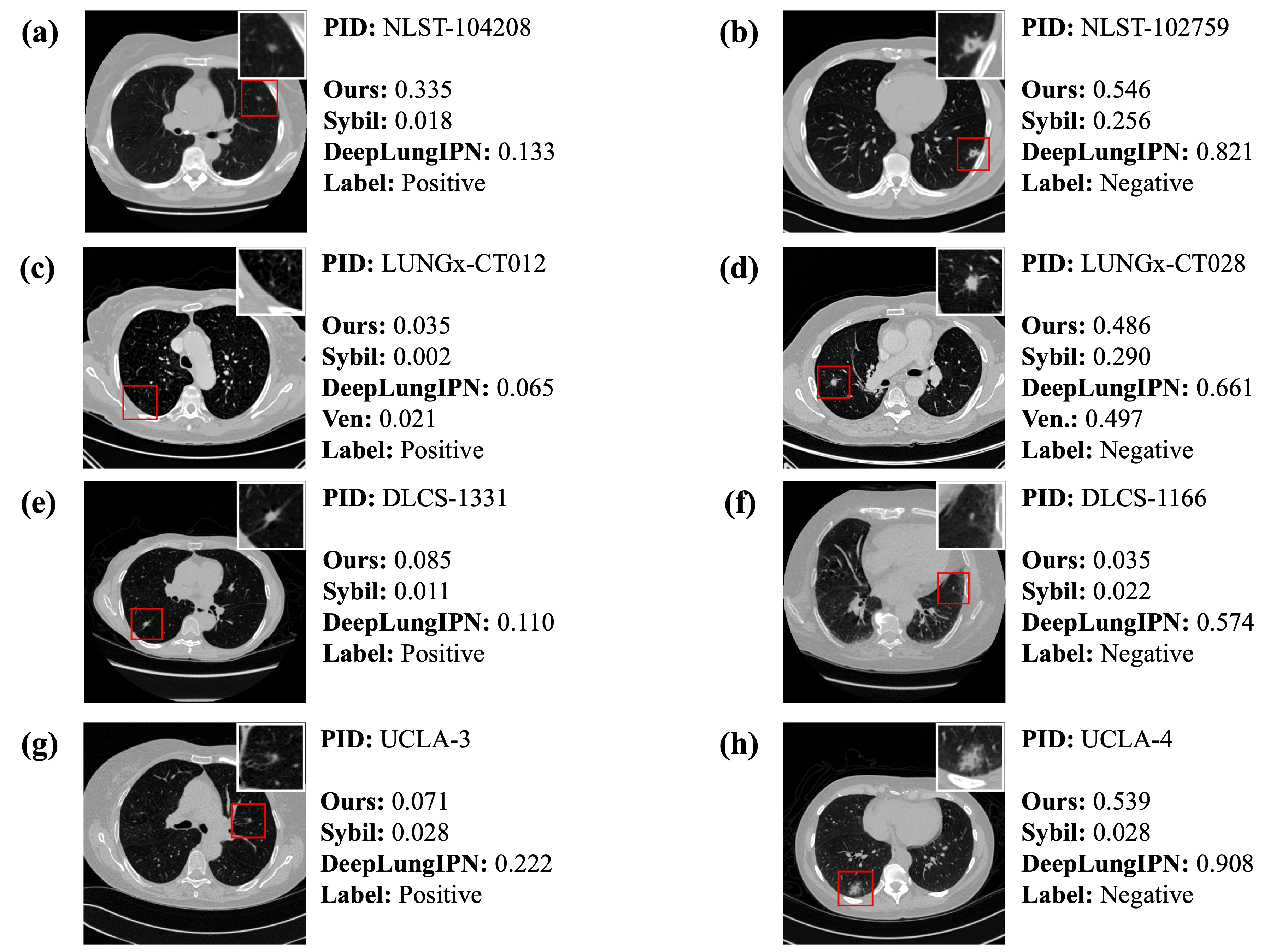}}
\caption{\textbf{Error Analysis (Extended).} One lung cancer and one non-lung cancer case from each dataset are presented. We present the CT scan slice corresponding to the middle of the nodule, highlighted with a red bounding box and magnified in the top right corner. Certain features may not be fully appreciable in the single slice but are visible when viewing the whole series. Patient-level risk scores are shown beside images. }
\label{fig3_appendix}
\end{figure}

Fig. \ref{fig3_appendix}c, e, and g depict nodules diagnosed as lung cancer, but our model underestimates the risk. Although these nodules are relatively small, they exhibit irregular shapes (c), spiculated margins (e), mixed densities (g), and pleural attachment (c, e, g). Interestingly, all other models also predict low scores for these nodules. In Fig. \ref{fig3_appendix}a, a lobulated, solid, well-marginated nodule is shown with a few suspicious features. It is reasonable that our model predicts a 33\% likelihood of malignancy, but all other models assign extremely low scores to this cancer patient.

In Fig. \ref{fig3_appendix}b, the nodule is solid with spiculated and lobulated margins and an irregular shape. Concerning characteristics include cyst-like spaces and pleural retraction. In Fig. \ref{fig3_appendix}d, the nodule appears to be solid with spiculated margins, septal stretching, and vascular convergence. In Fig. \ref{fig3_appendix}h, a large, part-solid nodule with poorly defined margins is observed. Given these suspicious features, most models assign relatively higher risk scores, despite the patient not being diagnosed with lung cancer. However, in Fig. \ref{fig3_appendix}h, Sybil accurately identifies it as having a lower score. In contrast, Fig. \ref{fig3_appendix}f shows a tiny nodule with no apparent suspicious features, but DeepLungIPN assigns a comparatively higher risk. We found that the nodules detected by DeepLungIPN appear to be fibrosis and vessels.

\nolinenumbers 
\FloatBarrier

\clearpage
\bibliographystyle{elsarticle-num} 
\bibliography{main}






\end{document}